\documentclass[10pt,twocolumn,letterpaper]{article}

\usepackage{iccv}
\usepackage{times}
\usepackage{epsfig}
\usepackage{graphicx}
\usepackage{amsmath}
\usepackage{amssymb}
\usepackage{paralist}
\usepackage{booktabs}
\usepackage{bbm}

\usepackage[pagebackref=true,breaklinks=true,letterpaper=true,colorlinks,bookmarks=false]{hyperref}

\iccvfinalcopy

\DeclareMathOperator*{\argmax}{\arg\!\max}

\newcommand{\est}[1]{\hat{#1}}
\newcommand{\gt}[1]{{#1}^*}

\newcommand{\crd}{\mathbf{y}}
\newcommand{\crds}{\mathcal{Y}}

\newcommand{\mdl}{\mathbf{h}}
\newcommand{\pool}{\mathcal{H}}
\newcommand{\loss}{\ell}
\newcommand{\Loss}{\mathcal{L}}

\newcommand{\param}{\mathbf{w}}

\newcommand{\expectation}[2]{\mathbb{E}_{#1}\left[ #2 \right]}
\newcommand{\derv}[1]{\frac{\partial}{\partial #1}}

\begin{document}

\title{\vspace{-0.4cm}Neural-Guided RANSAC: Learning Where to Sample Model Hypotheses}

\author{Eric Brachmann and Carsten Rother\\
Visual Learning Lab\\
Heidelberg University (HCI/IWR)\\
{\tt\small http://vislearn.de}
}

\maketitle

\begin{abstract}
\vspace{-0.4cm}
We present Neural-Guided RANSAC (NG-RANSAC), an extension to the classic RANSAC algorithm from robust optimization.
NG-RANSAC uses prior information to improve model hypothesis search, increasing the chance of finding outlier-free minimal sets.
Previous works use heuristic side information like hand-crafted descriptor distance to guide hypothesis search.
In contrast, we learn hypothesis search in a principled fashion that lets us optimize an arbitrary task loss during training, leading to large improvements on classic computer vision tasks.
We present two further extensions to NG-RANSAC.
Firstly, using the inlier count itself as training signal allows us to train neural guidance in a self-supervised fashion. 
Secondly, we combine neural guidance with differentiable RANSAC to build neural networks which focus on certain parts of the input data and make the output predictions as good as possible. 
We evaluate NG-RANSAC on a wide array of computer vision tasks, namely estimation of epipolar geometry, horizon line estimation and camera re-localization. 
We achieve superior or competitive results compared to state-of-the-art robust estimators, including very recent, learned ones.
\end{abstract}

\vspace{-0.4cm}
\section{Introduction}

\begin{figure}[t]
\begin{center}
   \vspace{-0.3cm}
   \includegraphics[width=1.0\linewidth]{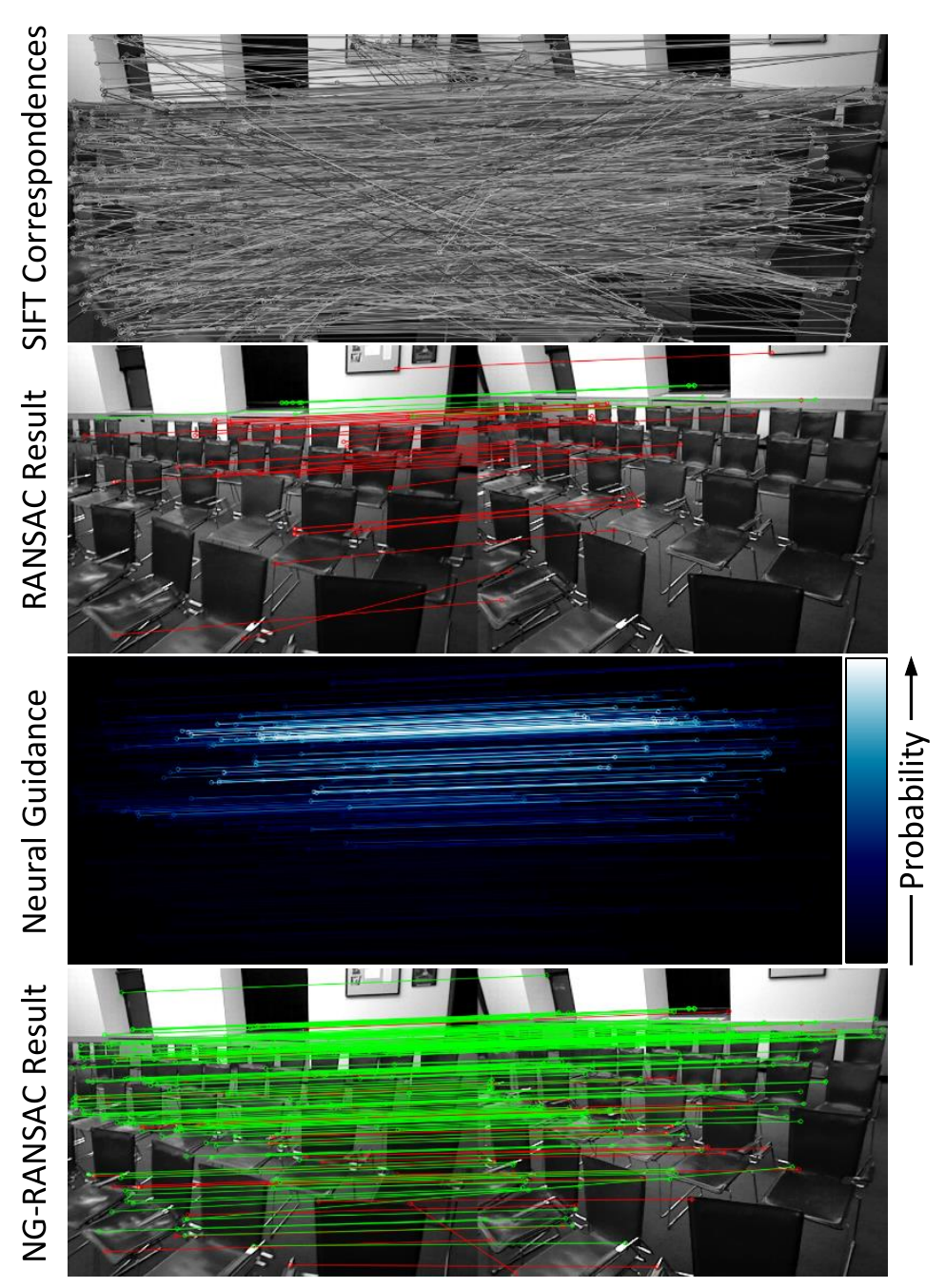}
\end{center}
   \vspace{-0.4cm}
   \caption{\textbf{RANSAC vs.~NG-RANSAC.} We extract 2000 SIFT correspondences between two images. 
   With an outlier rate of 88\%, RANSAC fails to find the correct relative transformation (green correct and red wrong matches). 
   We use a neural network to predict a probability distribution over correspondences. 
   Over 90\% of the probability mass falls onto 239 correspondences with an outlier rate of 33\%.
   NG-RANSAC samples minimal sets according to this distribution, and finds the correct transformation up to an angular error of less than 1$^\circ$.
   }
   \vspace{-0.3cm}
\label{fig:teaser}
\end{figure}

Despite its simplicity and time of invention, Random Sample Consensus (RANSAC) \cite{ransac1981} remains an important method for robust optimization, and is a vital component of many state-of-the-art vision pipelines \cite{sattler2016efficient, schoenberger2016sfm, orbslam22017, brachmann2018lessmore}.
RANSAC allows accurate estimation of model parameters from a set of observations of which some are outliers.
To this end, RANSAC iteratively chooses random sub-sets of observations, so called minimal sets, to create model hypotheses.
Hypotheses are ranked according to their consensus with all observations, and the top-ranked hypothesis is returned as the final estimate.

The main limitation of RANSAC is its poor performance in domains with many outliers.
As the ratio of outliers increases, RANSAC requires exponentially many iterations to find an outlier-free minimal set.
Implementations of RANSAC therefore often restrict the maximum number of iterations, and return the best model found so far \cite{opencv_library}.

In this work, we combine RANSAC with a neural network that predicts a weight for each observation.
The weights ultimately guide the sampling of minimal sets.
We call the resulting algorithm Neural-Guided RANSAC (NG-RANSAC).
A comparison of our method with vanilla RANSAC can be seen in Fig.~\ref{fig:teaser}.

When developing NG-RANSAC, we took inspiration from recent work on learned robust estimators \cite{goodcorr18, deepfund18}.
In particular, Yi \etal \cite{goodcorr18} train a neural network to classify observations as outliers or inliers, fitting final model parameters only to the latter.
Although designed to replace RANSAC, their method achieves best results when combined with RANSAC during test time, where it would remove any outliers that the neural network might have missed.
This motivates us to train the neural network in conjunction with RANSAC in a principled fashion, rather than imposing it afterwards.

Instead of interpreting the neural network output as soft inlier labels for a robust model fit, we let the output weights guide RANSAC hypothesis sampling.
Intuitively, the neural network should learn to decrease weights for outliers, and increase them for inliers.
This paradigm yields substantial flexibility for the neural network in allowing a certain misclassification rate without negative effects on the final fitting accuracy due to the robustness of RANSAC.
The distinction between inliers and outliers, as well as which misclassifications are tolerable, is solely guided by the minimization of the task loss function during training.
Furthermore, our formulation of NG-RANSAC facilitates training with any (non-differentiable) task loss function, and any (non-differentiable) model parameter solver, making it broadly applicable.
For example, when fitting essential matrices, we may use the 5-point algorithm rather than the (differentiable) 8-point algorithm which other learned robust estimators rely on \cite{goodcorr18, deepfund18}.
The flexibility in choosing the task loss also allows us to train NG-RANSAC self-supervised by using maximization of the inlier count as training objective.

The idea of using guided sampling in RANSAC is not new. 
Tordoff and Murray first proposed to guide the hypothesis search of MLESAC \cite{mlesac2000}, using side information \cite{guidedmlesac2002}.
They formulated a prior probability of sparse feature matches being valid based on matching scores. 
While this has a positive affect on RANSAC performance in some applications, feature matching scores, or other hand-crafted heuristics, were clearly \emph{not designed} to guide hypothesis search.
In particular, calibration of such ad-hoc measures can be difficult as the reliance on over-confident but wrong prior probabilities can yield situations where the same few observations are sampled repeatedly.
This fact was recognized by Chum and Matas who proposed PROSAC \cite{prosac2005}, a variant of RANSAC that uses side information only to change \emph{the order} in which RANSAC draws minimal sets.
In the worst case, if the side information was not useful at all, their method would degenerate to vanilla RANSAC.
NG-RANSAC takes a different approach in (i) learning the weights to guide hypothesis search rather than using hand-crafted heuristics, and (ii) integrating RANSAC itself in the training process which leads to self-calibration of the predicted weights.

Recently, Brachmann \etal proposed differentiable RANSAC (DSAC) to learn a camera re-localization pipeline \cite{brachmann2017dsac}.
Unfortunately, we can not directly use DSAC to learn hypothesis sampling since DSAC is only differentiable \wrt to observations, not sampling weights.
However, NG-RANSAC applies a similar trick also used to make DSAC differentiable, namely the optimization of the expected task loss during training.
While we do not rely on DSAC, neural guidance can be used in conjunction with DSAC (NG-DSAC) to train neural networks that predict observations and observation confidences at the same time.

\vspace{0.1cm}
\noindent We summarize our main contributions:
\vspace{0.1cm}
\begin{compactitem}
\item We present NG-RANSAC, a formulation of RANSAC with learned guidance of hypothesis sampling. We can use any (non-differentiable) task loss, and any (non-differentiable) minimal solver for training.
\item Choosing the inlier count itself as training objective facilitates self-supervised learning of NG-RANSAC.
\item We use NG-RANSAC to estimate epipolar geometry of image pairs from sparse correspondences, where it surpasses competing robust estimators.
\item We combine neural guidance with differentiable RANSAC (NG-DSAC) to train neural networks that make accurate predictions for parts of the input, while neglecting other parts. These models achieve competitive results for horizontal line estimation, and state-for-the-art for camera re-localization.
\end{compactitem}

\vspace{-0.1cm}
\section{Related Work}
\label{sec:related}
\vspace{-0.1cm}

RANSAC was introduced in 1981 by Fischler and Bolles \cite{ransac1981}. 
Since then it was extended in various ways, see \eg the survey by Raguram \etal \cite{ransacsurvey2008}.
Combining some of the most promising improvements, Raguram \etal created the Universal RANSAC (USAC) framework \cite{usac2013} which represents the state-of-the-art of classic RANSAC variants.
USAC includes guided hypothesis sampling according to PROSAC \cite{prosac2005}, more accurate model fitting according to Locally Optimized RANSAC \cite{Chum2003}, and more efficient hypothesis verification according to Optimal Randomized RANSAC \cite{orransac2008}.
Many of the improvements proposed for RANSAC could also be applied to NG-RANSAC since we do not require any differentiability of such add-ons. 
We only impose restrictions on how to generate hypotheses, namely according to a learned probability distribution.

RANSAC is not often used in recent machine learning-heavy vision pipelines. 
Notable exceptions include geometric problems like object instance pose estimation \cite{brachmann2014pose6d, brachmann2016, ipose2018}, and camera re-localization \cite{shotton13scorf,valentin2015cvpr,rfvscnn2016,cavallari2017fly,taira2018inloc} where RANSAC is coupled with decision forests or neural networks that predict image-to-object correspondences.
However, in most of these works, RANSAC is not part of the training process because of its non-differentiability.
DSAC \cite{brachmann2017dsac, brachmann2018lessmore} overcomes this limitation by making the hypothesis selection a probabilistic action which facilitates optimization of the expected task loss during training.
However, DSAC is limited in \emph{which} derivatives can be calculated.
DSAC allows differentiation \wrt to observations.
For example, we can use it to calculate the gradient of image coordinates for a sparse correspondence.
However, DSAC does not model observation selection, and hence we cannot use it to optimize a matching probability.
By showing how to learn neural guidance, we close this gap. 
The combination with DSAC enables the full flexibility of learning both, observations and their selection probability.

Besides DSAC, a \emph{differentiable} robust estimator, there has recently been some work on \emph{learning} robust estimators.
We discussed the work of Yi \etal \cite{goodcorr18} in the introduction.
Ranftl and Koltun \cite{deepfund18} take a similar but iterative approach reminiscent of Iteratively Reweighted Least Squares (IRLS) for fundamental matrix estimation.
In each iteration, a neural network predicts observation weights for a weighted model fit, taking into account the residuals of the last iteration.
Both, \cite{goodcorr18} and \cite{deepfund18}, have shown considerable improvements \wrt to vanilla RANSAC but require differentiable minimal solvers, and task loss functions.
NG-RANSAC outperforms both approaches, and is more flexible when it comes to defining the training objective.
This flexibility also enables us to train NG-RANSAC in a self-supervised fashion, possible with neither \cite{goodcorr18} nor \cite{deepfund18}.

\section{Method}
\label{sec:method}

\noindent \textbf{Preliminaries.}
We address the problem of fitting model parameters $\mdl$ to a set of observations $\crd \in \crds$ that are contaminated by noise and outliers.
For example, $\mdl$ could be a fundamental matrix that describes the epipolar geometry of an image pair \cite{mutliview2004}, and $\crds$ could be the set of SIFT correspondences \cite{Lowesift} we extract for the image pair.
To calculate model parameters from the observations, we utilize a solver $f$, for example the 8-point algorithm \cite{eightpoint97}.
However, calculating $\mdl$ from all observations will result in a poor estimate due to outliers.
Instead, we can calculate $\mdl$ from a small subset (minimal set) of observations with cardinality $N$: $\mdl = f(\crd_1, \dots, \crd_N)$.
For example, for a fundamental matrix $N=8$ when using the 8-point algorithm.
RANSAC \cite{ransac1981} is an algorithm to chose an outlier-free minimal set from $\crds$ such that the resulting estimate $\mdl$ is accurate.
To this end, RANSAC randomly chooses $M$ minimal sets to create a pool of model hypotheses $\pool = (\mdl_1, \dots, \mdl_M)$.

RANSAC includes a strategy to adaptively choose $M$, based on an online estimate of the outlier ratio \cite{ransac1981}.
The strategy guarantees that an outlier-free set will be sampled with a user-defined probability.
For tasks with large outlier ratios, $M$ calculated like this can be exponentially large, and is usually clamped to a maximum value \cite{opencv_library}.
For notational simplicity, we take the perspective of a fixed $M$ but do not restrict the use of an early-stopping strategy in practice.

RANSAC chooses a model hypothesis as the final estimate $\est{\mdl}$ according to a scoring function $s$:
\begin{equation}
\label{eq:ransac}
\est{\mdl} = \argmax_{\mdl \in \pool} s(\mdl, \crds).
\end{equation}
The scoring function measures the consensus of an hypothesis \wrt all observations, and is traditionally implemented as inlier counting \cite{ransac1981}.

\noindent \textbf{Neural Guidance.}
RANSAC chooses observations uniformly random to create the hypothesis pool $\pool$. 
We aim at sampling observations according to a learned distribution instead that is parametrized by a neural network with parameters $\param$.
That is, we select observations according to $\crd \sim p(\crd;\param)$.
Note that $p(\crd;\param)$ is a categorical distribution over the discrete set of observations $\crds$, \emph{not} a continuous distribution in observation space.
We wish to learn parameters $\param$ in a way that increases the chance of selecting outlier-free minimal sets, which will result in accurate estimates $\est{\mdl}$.
We sample a hypothesis pool $\pool$ according to $p(\pool;\param)$ by sampling observations and minimal sets independently, \ie
\begin{gather}
\label{eq:probpool}
p(\pool;\param) = \prod_{j=1}^M~p(\mdl_j;\param),~\text{with}~~p(\mdl;\param) = \prod_{i=1}^N~p(\crd_i;\param).
\end{gather}
From a pool $\pool$, we estimate model parameters $\est{\mdl}$ with RANSAC according to Eq.~\ref{eq:ransac}.
For training, we assume that we can measure the quality of the estimate with a task loss function $\loss(\est{\mdl})$.
The task loss can be calculated \wrt a ground truth model $\gt{\mdl}$, or self-supervised, \eg by using the inlier count of the final estimate: $\loss(\est{\mdl}) = -s(\est{\mdl}, \crds)$.
We wish to learn the distribution $p(\pool;\param)$ in a way that we receive a small task loss with high probability.
Inspired by DSAC \cite{brachmann2017dsac}, we define our training objective as the minimization of the expected task loss:
\begin{equation}
\label{eq:ng-ransac}
\Loss(\param) = \expectation{\pool \sim p(\pool;\param)}{\loss(\est{\mdl})}.
\end{equation}
We compute the gradients of the expected task loss \wrt the network parameters as
\begin{gather}
\derv{\param} \Loss(\param) = \expectation{\pool}{\loss(\est{\mdl}) ~ \derv{\param} \log p(\pool;\param)}.
\end{gather}
Integrating over all possible hypothesis pools to calculate the expectation is infeasible. 
Therefore, we approximate the gradients by drawing $K$ samples $\pool_k \sim p(\pool;\param)$:
\begin{gather}
\label{eq:ng-ransac-grad}
\derv{\param} \Loss(\param) \approx \frac{1}{K} \sum_{k=1}^K \left[ \loss(\est{\mdl}) \derv{\param} \log p(\pool_k;\param) \right].
\end{gather}
Note that gradients of the task loss function $\loss$ do \emph{not} appear in the expression above. 
Therefore, differentiability of the task loss $\loss$, the robust solver $\est{\mdl}$ (\ie RANSAC) or the minimal solver $f$ is not required.
These components merely generate a training signal for steering the sampling probability $p(\pool;\param)$ in a good direction.
Due to the approximation by sampling, the gradient variance of Eq.~\ref{eq:ng-ransac-grad} can be high.
We apply a standard variance reduction technique from reinforcement learning by subtracting a baseline $b$ \cite{rlbook}:
\begin{gather}
\label{eq:ng-ransac-grad-final}
\derv{\param} \Loss(\param) \approx \frac{1}{K} \sum_{k=1}^K \left[ [\loss(\est{\mdl}) - b] ~ \derv{\param} \log p(\pool_k;\param) \right].
\end{gather}
We found a simple baseline in the form of the average loss per image sufficient, \ie $b = \bar{\loss}$.
Subtracting the baseline will move the probability distribution towards hypothesis pools with lower-than-average loss for each training example.

\noindent \textbf{Combination with DSAC.}
Brachmann \etal \cite{brachmann2017dsac} proposed a RANSAC-based pipeline where a neural network with parameters $\param$ predicts observations $\crd(\param) \in \crds(\param)$.
End-to-end training of the pipeline, and therefore learning the observations $\crd(\param)$, is possible by turning the $\argmax$ hypothesis selection of RANSAC (cf.~Eq.~\ref{eq:ransac}) into a probabilistic action:
\begin{equation}
\label{eq:dsac}
\est{\mdl}_\text{DSAC} = \mdl_j \sim p(j|\pool) = \frac{\exp s(\mdl_j, \crds(\param))}{\sum_{k=1}^M \exp s(\mdl_k, \crds(\param))}.
\end{equation}
This differentiable variant of RANSAC (DSAC) chooses a hypothesis randomly according to a distribution calculated from hypothesis scores. 
The training objective aims at learning network parameters such that hypotheses with low task loss are chosen with high probability:
\begin{equation}
\label{eq:dsac-train}
\Loss_\text{DSAC}(\param) =\expectation{j \sim p(j)}{\loss(\mdl_j)}.
\end{equation}
In the following, we extend the formulation of DSAC with neural guidance (NG-DSAC). 
We let the neural network predict observations $\crd(\param)$ and, additionally, a probability associated with each observation $p(\crd;\param)$.
Intuitively, the neural network can express a confidence in its own predictions through this probability. 
This can be useful if a certain input for the neural network contains no information about the desired model $\mdl$.
In this case, the observation prediction $\crd(\param)$ is necessarily an outlier, and the best the neural network can do is to label it as such by assigning a low probability.
We combine the training objectives of NG-RANSAC (Eq.~\ref{eq:ng-ransac}) and DSAC (Eq.~\ref{eq:dsac-train}) which yields:
\begin{equation}
\label{eq:ng-dsac}
\Loss_\text{NG-DSAC}(\param) = \mathbb{E}_{\pool \sim p(\pool;\param)}\expectation{j \sim p(j|\pool)}{\loss(\mdl_j)},
\end{equation}
where we again construct $p(\pool;\param)$ from individual $p(\crd;\param)$'s according to Eq.~\ref{eq:probpool}.
The training objective of NG-DSAC consists of two expectations.
Firstly, the expectation \wrt sampling a hypothesis pool according to the probabilities predicted by the neural network.
Secondly, the expectation \wrt sampling a final estimate from the pool according to the scoring function.
As in NG-RANSAC, we approximate the first expectation via sampling, as integrating over all possible hypothesis pools is infeasible.
For the second expectation, we can calculate it analytically, as in DSAC, since it integrates over the discrete set of hypotheses $\mdl_j$ in a given pool $\pool$.
Similar to Eq.~\ref{eq:ng-ransac-grad-final}, we give the approximate gradients $\derv{\param} \Loss(\param)$ of NG-DSAC as:
\begin{gather}
\label{eq:ng-dsac-grad}
\frac{1}{K} \sum_{k=1}^K \left[ [\expectation{j}{\loss} - b] ~ \derv{\param} \log p(\pool_k;\param) + \derv{\param} \expectation{j}{\loss}\right],
\end{gather}
where we use $\expectation{j}{\loss}$ as a stand-in for $\expectation{j \sim p(j|\pool_k)}{\loss(\mdl_j)}$.
The calculation of gradients for NG-DSAC requires the derivative of the task loss (note the last part of Eq.~\ref{eq:ng-dsac-grad}) because $\expectation{j}{\loss}$ depends on parameters $\param$ via observations $\crd(\param)$.
Therefore, training NG-DSAC requires a differentiable task loss function $\loss$, a differentiable scoring function $s$, and a differentiable minimal solver $f$. 
Note that we inherit these restrictions from DSAC.
In return, NG-DSAC allows for learning observations and observation confidences, at the same time.

\section{Experiments}
\label{sec:experiments}

We evaluate neural guidance on multiple, classic computer vision tasks.
Firstly, we apply NG-RANSAC to estimating epipolar geometry of image pairs in the form of essential matrices and fundamental matrices.
Secondly, we apply NG-DSAC to horizon line estimation and camera re-localization.
We present the main experimental results here, and refer to the appendix for details about network architectures, hyper-parameters and more qualitative results.
Our implementation is based on PyTorch \cite{paszke2017automatic}, and we will make the code publicly available for all tasks discussed below\footnote{\url{vislearn.de/research/neural-guided-ransac/}}.

\begin{figure*}[h!]
\begin{center}
   \includegraphics[width=1.0\linewidth]{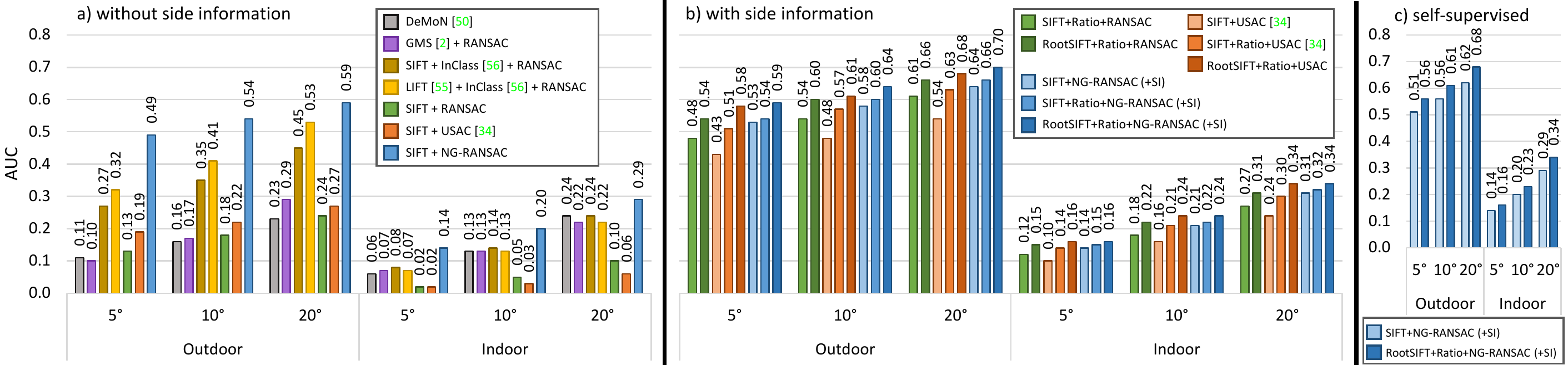}
\end{center}
\vspace{-0.3cm}
   \caption{\textbf{Essential Matrix Estimation.} We calculate the relative pose between outdoor and indoor image pairs via the essential matrix. We measure the AUC of the cumulative angular error up to a threshold of 5$^\circ$, 10$^\circ$ or 20$^\circ$. \textbf{a)} We use no side information about the sparse correspondences. \textbf{b)} We use side information in the form of descriptor distance ratios between the best and second best match. We use it to filter correspondences with a threshold of 0.8 (\emph{+Ratio}), as an additional input for our network (\emph{+SI}), and as additional input for \mbox{USAC \cite{usac2013}}. \textbf{c)} We train NG-RANSAC in a self-supervised fashion by using the inlier count as training objective.}
   \label{fig:goodcorr}
   \vspace{-0.3cm}
\end{figure*}

\subsection{Essential Matrix Estimation}
\label{sec:exp:emat}

Epipolar geometry describes the geometry of two images that observe the same scene \cite{mutliview2004}.
In particular, two image points $\mathbf{x}$ and $\mathbf{x}'$ in the left and right image corresponding to the same 3D point satisfy $\mathbf{x}'^\top F \mathbf{x}=0$, where the $3 \times 3$ matrix $F$ denotes the fundamental matrix.
We can estimate $F$ uniquely (but only up to scale) from 8 correspondences, or from 7 correspondences with multiple solutions \cite{mutliview2004}.
The essential matrix $E$ is a special case of the fundamental matrix when the calibration parameters $K$ and $K'$ of both cameras are known: $E=K'^\top F K$.
The essential matrix can be estimated from 5 correspondences \cite{fivepoint04}.
Decomposing the essential matrix allows to recover the relative pose between the observing cameras, and is a central step in image-based 3D reconstruction \cite{schoenberger2016sfm}.
As such, estimating the fundamental or essential matrices of image pairs is a classic and well-researched problem in computer vision.

In the following, we firstly evaluate NG-RANSAC for the calibrated case and estimate essential matrices from SIFT correspondences \cite{Lowesift}.
For the sake of comparability with the recent, learned robust estimator of Yi \etal \cite{goodcorr18} we adhere closely to their evaluation setup, and compare to their results.

\noindent \textbf{Datasets.} 
Yi \etal \cite{goodcorr18} evaluate their approach in outdoor as well as indoor settings.
For the outdoor datasets, they select five scenes from the structure-from-motion (SfM) dataset of \cite{reconworld15}: \emph{Buckingham}, \emph{Notredame}, \emph{Sacre Coeur}, \emph{St. Peter's} and \emph{Reichstag}.
They pick two additional scenes from \cite{benchmvs08}: \emph{Fountain} and \emph{Herzjesu}.
They reconstruct each scene using a SfM tool \cite{visualsfm13} to obtain `ground truth' camera poses, and co-visibility constraints for selecting image pairs.
For indoor scenes Yi \etal choose 16 sequences from the SUN3D dataset \cite{sun3d13} which readily comes with ground truth poses captured by KinectFusion \cite{newcombe2011kinectfusion}.
See Appendix \ref{app:emat} for a listing of all scenes.
Indoor scenarios are typically very challenging for sparse feature-based approaches because of texture-less surfaces and repetitive elements (see Fig.~\ref{fig:teaser} for an example).
Yi \etal train their best model using one outdoor scene (\emph{St. Peter's}) and one indoor scene (\emph{Brown 1}), and test on all remaining sequences (6 outdoor, 15 indoor).
Yi \etal kindly provided us with their exact data splits, and we will use their setup.
Note that training and test is performed on completely separate scenes, \ie the neural network has to generalize to unknown environments.

\noindent \textbf{Evaluation Metric.}
Via the essential matrix, we recover the relative camera pose up to scale, and compare to the ground truth pose as follows.
We measure the angular error between the pose rotations, as well as the angular error between the pose translation vectors in degrees.
We take the maximum of the two values as the final angular error.
We calculate the cumulative error curve for each test sequence, and compute the area under the curve (AUC) up to a threshold of 5$^\circ$, 10$^\circ$ or 20$^\circ$.
Finally, we report the average AUC over all test sequences (but separately for the indoor and outdoor setting).

\noindent \textbf{Implementation}.
Yi \etal train a neural network to classify a set of sparse correspondences in inliers and outliers.
They represent each correspondence as a 4D vector combining the 2D coordinate in the left and right image.
Their network is inspired by PointNet \cite{qi2016pointnet}, and processes each correspondence independently by a series of multilayer perceptrons (MLPs). 
Global context is infused by using instance normalization \cite{Ulyanov2016InstanceNT} in-between layers.
We re-build this architecture in PyTorch, and train it according to NG-RANSAC (Eq.~\ref{eq:ng-ransac}).
That is, the network predicts weights to guide RANSAC sampling instead of inlier class labels.
We use the angular error between the estimated relative pose, and the ground truth pose as task loss $\loss$.
As minimal solver $f$, we use the 5-point algorithm \cite{fivepoint04}.
To speed up training, we initialize the network by learning to predict the distance of each correspondence to the ground truth epipolar line, see Appendix \ref{app:emat} for details.
We initialize for 75k iterations, and train according to Eq.~\ref{eq:ng-ransac} for 25k iterations.
We optimize using Adam \cite{adam2014} with a learning rate of 10$^{-5}$.
For each training image, we extract 2000 SIFT correspondences, and sample $K=4$ hypothesis pools with $M=16$ hypotheses. 
We use a low number of hypotheses during training to obtain variation when sampling pools.
For testing, we increase the number of hypotheses to $M=10^3$.
We use an inlier threshold of 10$^{-3}$ assuming normalized image coordinates using camera calibration parameters.

\begin{figure*}[t]
\vspace{-0.3cm}
\begin{center}
   \includegraphics[width=1.0\linewidth]{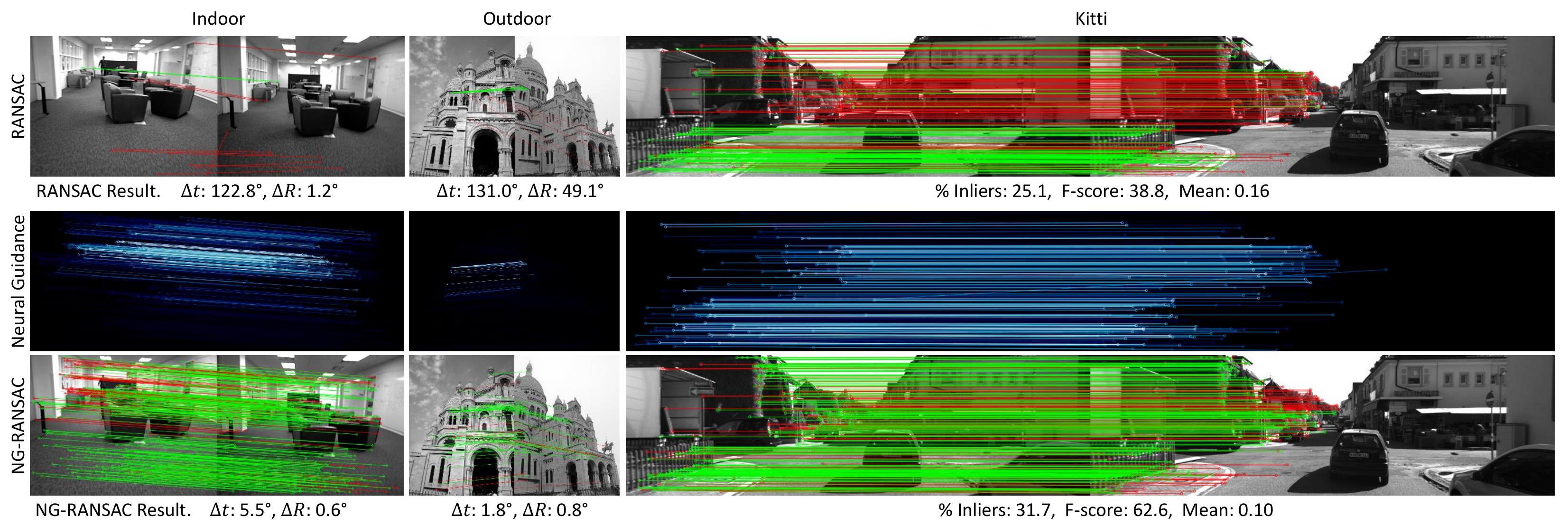}
\end{center}
\vspace{-0.4cm}
   \caption{\textbf{Qualitative Results.} We compare fitted models for RANSAC and NG-RANSAC. For the indoor and outdoor image pairs, we fit essential matrices, and for the Kitti image pair we fit the fundamental matrix. We draw final model inliers in green if they adhere to the ground truth model, and red otherwise. We also measure the quality of each estimate, see the main text for details on the metrics.}
   \vspace{-0.4cm}
   \label{fig:goodcorr2}
\end{figure*}

\noindent \textbf{Results.}
We compare NG-RANSAC to the inlier classification (\emph{InClass}) of Yi \etal \cite{goodcorr18}.
They use their approach with SIFT as well as LIFT \cite{lift2016} features.
We include results for DeMoN \cite{demon17}, a learned SfM pipeline, and GMS \cite{bian2017gms}, a semi-dense approach using ORB features \cite{rublee2011orb}.
As classical baselines, we compare to vanilla RANSAC \cite{ransac1981} and USAC \cite{usac2013}.
See Fig.~\ref{fig:goodcorr} a) for results. 
RANSAC achieves poor results for indoor and outdoor scenes across all thresholds, scoring as the weakest method.
In this experiment, we assume no side information is available about the quality of correspondences.
Therefore, USAC  performs similar to RANSAC, since it cannot use guided sampling.
Coupling RANSAC with neural guidance (NG-RANSAC) elevates it to the leading position with a comfortable margin.
Different from USAC, NG-RANSAC deduces useful guiding weights solely from the spatial distribution of correspondences.
See also Fig.~\ref{fig:goodcorr2} for qualitative results.

NG-RANSAC outperforms \emph{InClass} of Yi \etal \cite{goodcorr18} despite some similarities.
Both use the same network architecture, are based on SIFT correspondences, and both use RANSAC at test time.
Yi \etal \cite{goodcorr18} train using a hybrid classification-regression loss based on the 8-point algorithm, and ultimately compare essential matrices using squared error.
Therefore, their training objective is very different from the evaluation procedure.
During evaluation, they use RANSAC with the 5-point algorithm on top of their inlier predictions, and measure the angular error.
NG-RANSAC incorporates all these components in its training procedure, and therefore optimizes the correct objective.

\noindent \textbf{Using Side Information.}
The evaluation procedure of Yi \etal \cite{goodcorr18} is designed to test a robust estimator in high-outlier domains.
However, it underestimates what classical approaches can achieve on these datasets.
The distance ratio of the best and second-best SIFT match is often an indicator of correspondence quality.
This side information can be used by USAC \cite{usac2013} to guide hypothesis sampling according to the PROSAC strategy \cite{prosac2005}.
Furthermore, Lowe's ratio criterion \cite{Lowesift} removes ambiguous matches with a distance ratio above a threshold (we use 0.8) before running RANSAC.
We denote the ratio filter as \emph{+Ratio} in Fig.~\ref{fig:goodcorr} b), and observe a drastic improvement for all methods.
Both classic approaches, RANSAC and USAC, outperform all learned methods of Fig.~\ref{fig:goodcorr} a).
RootSIFT normalization of SIFT descriptors \cite{rootsift12} improves accuracy further.
NG-RANSAC easily incorporates side information.
For best accuracy, we train it on ratio-filtered RootSIFT correspondences, using distance ratios as additional network input (denoted by \emph{+SI}).

\begin{figure}[t]
\begin{center}
   \includegraphics[width=1.0\linewidth]{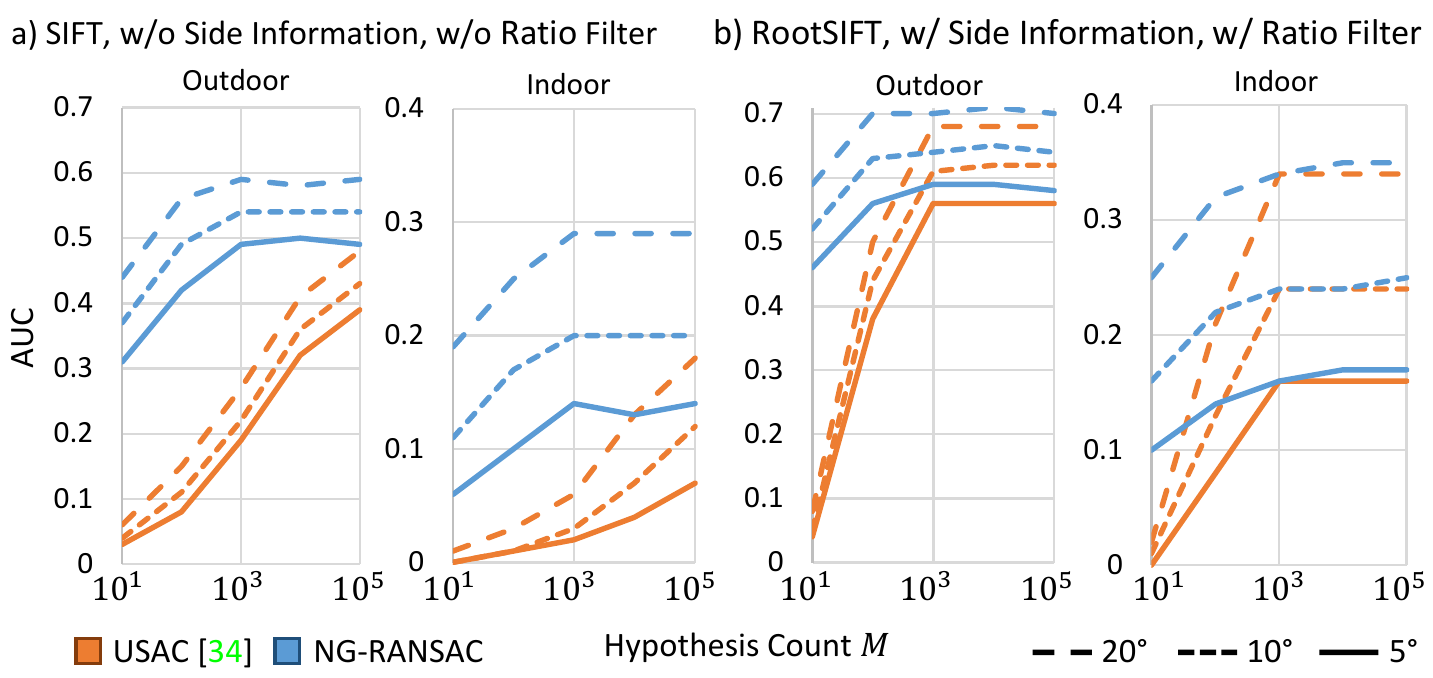}
\end{center}
\vspace{-0.4cm}
   \caption{\textbf{Accuracy vs.~Hypothesis Budget.} We compare the AUC of NG-RANSAC and USAC \cite{usac2013} for increasing number of hypotheses $M$. \textbf{a)} with and \textbf{b)} without side information. } 
   \label{fig:usac}
   \vspace{-0.4cm}
\end{figure}

\begin{figure*}[t]
\begin{center}
   \includegraphics[width=1.0\linewidth]{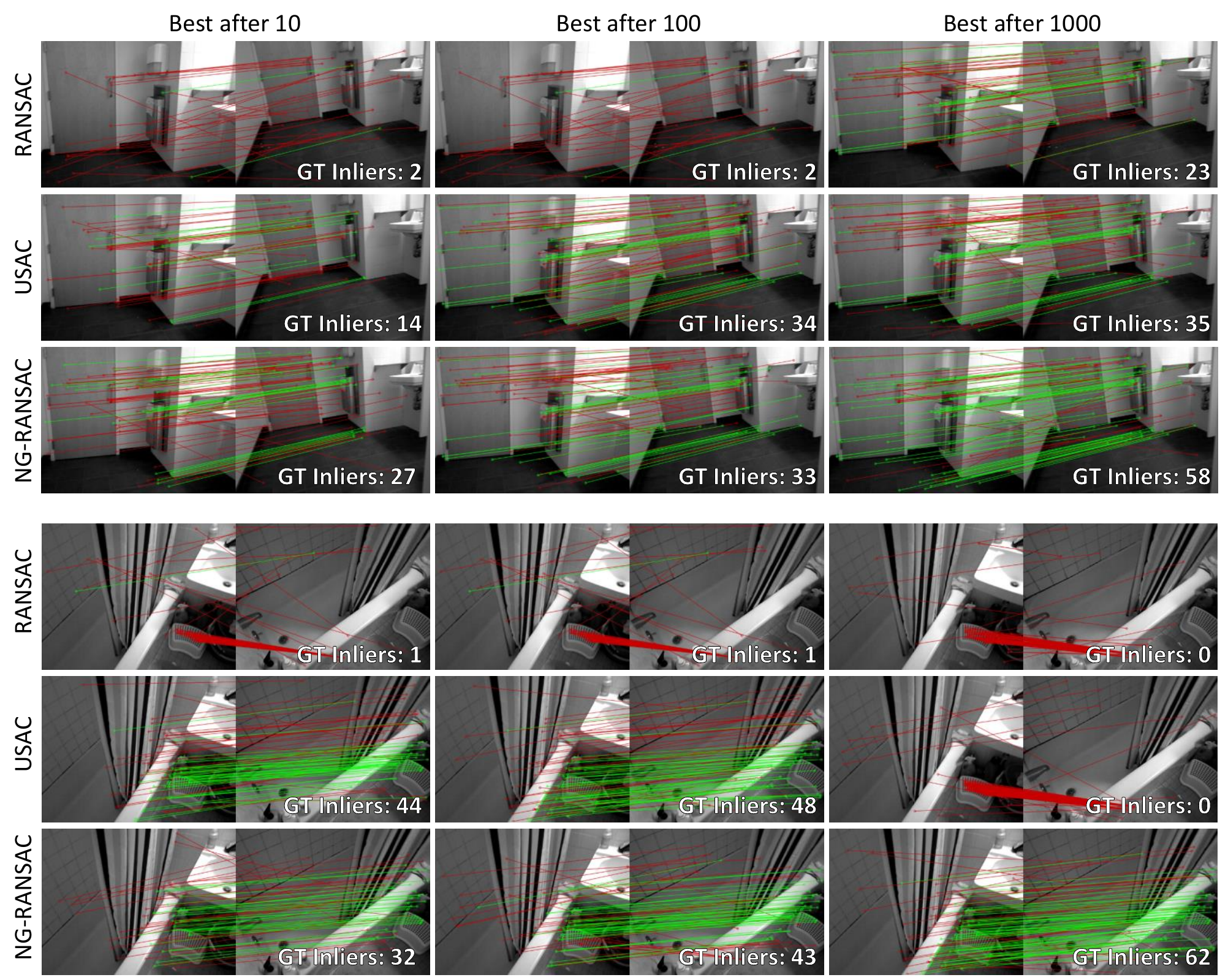}
\end{center}
\vspace{-0.4cm}
   \caption{\textbf{Hypothesis Search.} We visualize the best hypothesis found after $M \in \{10, 100, 1000\}$ iterations for RANSAC \cite{ransac1981}, USAC \cite{usac2013} and NG-RANSAC. For each result, we give the number of correspondences which are also inliers for the ground truth model (\emph{GT Inliers}, drawn in green). We perform this experiment in the \emph{Indoor} scenario, using side information and  RootSIFT but without Lowe's ratio filter.}
   \label{fig:usacqual}
   \vspace{-0.4cm}
\end{figure*}

The accuracy of USAC \cite{usac2013} and NG-RANSAC depend on the hypothesis budget $M$, see Fig.~\ref{fig:usac}.
NG-RANSAC finds good hypotheses much earlier than USAC, and achieves a reasonable accuracy by drawing as few as 10 hypotheses. 
Fig.~\ref{fig:usacqual} shows a visualization of progressive hypotheses search.
USAC is designed to draw the same hypotheses as RANSAC but in a different order.
Therefore, USAC samples degenerate hypotheses (poor accuracy but high inlier count) eventually, even if it gives them a low priority at first, see Fig.~\ref{fig:usacqual} bottom.
NG-RANSAC learns to suppress such hypotheses more effectively.

Interestingly, passing our learned weights to USAC achieves significantly lower accuracy than passing matching ratios to USAC.
For example, for the outdoor setting, w/o ratio filter and $M=10^3$, USAC achieves \mbox{-0.27/-0.24/-0.34} AUC for $5^\circ$/$10^\circ$/$20^\circ$ when using our weights.
The USAC/PROSAC sampling scheme assumes that the probability of correspondences being inliers increases monotonically with the sampling weight \cite{prosac2005}.
In contrast, our training objective optimizes over entire pools of hypotheses where correspondences are sampled independently.
Individual outlier correspondences might be ranked high by the neural network, without affecting accuracy negatively, thus violating the assumption of PROSAC.

\noindent \textbf{Self-supervised Learning.}
We train NG-RANSAC self-supervised by defining a task loss $\loss$ to assess the quality of an estimate independent of a ground truth model $\gt{\mdl}$.
A natural choice is the inlier count of the final estimate.
We found the inlier count to be a very stable training signal, even in the beginning of training such that we require no special initialization of the network.
We report results of self-supervised NG-RANSAC in Fig.~\ref{fig:goodcorr} c).
It outperforms all competitors except USAC \cite{usac2013} which it matches in accuracy. 
Unsupervised NG-RANSAC achieves slightly worse accuracy than supervised NG-RANSAC.
A supervised task loss allows NG-RANSAC to adapt more precisely to the evaluation measure used at test time.
For the datasets used so far, the process of image pairing uses co-visibility information, and therefore a form of supervision.
In the next section, we learn NG-RANSAC fully self-supervised by using the ordering of sequential data to assemble image pairs.

\noindent \textbf{Runtime.}
A forward pass of the network takes 3ms on CPU (similar for GPU). 
The total runtime (and accuracy) depends on the hypothesis count $M$. 
For $M=10^3$, our implementation of NG-RANSAC takes 90ms per image pair.
For $M=10$, it takes 21ms.

\subsection{Fundamental Matrix Estimation}

We apply NG-RANSAC to fundamental matrix estimation, comparing it to the learned robust estimator of Ranftl and Koltun \cite{deepfund18}, denoted \emph{Deep F-Mat}.
They propose an iterative procedure where a neural network estimates observation weights for a robust  model fit.
The residuals of the last iteration are an additional input to the network in the next iteration.
The network architecture is similar to the one used in \cite{goodcorr18}.
Correspondences are represented as 4D vectors, and they use the descriptor matching ratio as an additional input.
Each observation is processed by a series of MLPs with instance normalization interleaved.
\emph{Deep F-Mat} was published very recently, and the code is not yet available.
We therefore follow the evaluation procedure described in \cite{deepfund18} and compare to their results.

\begin{figure}[t]
\begin{center}
   \vspace{-0.3cm}
   \includegraphics[width=0.9\linewidth]{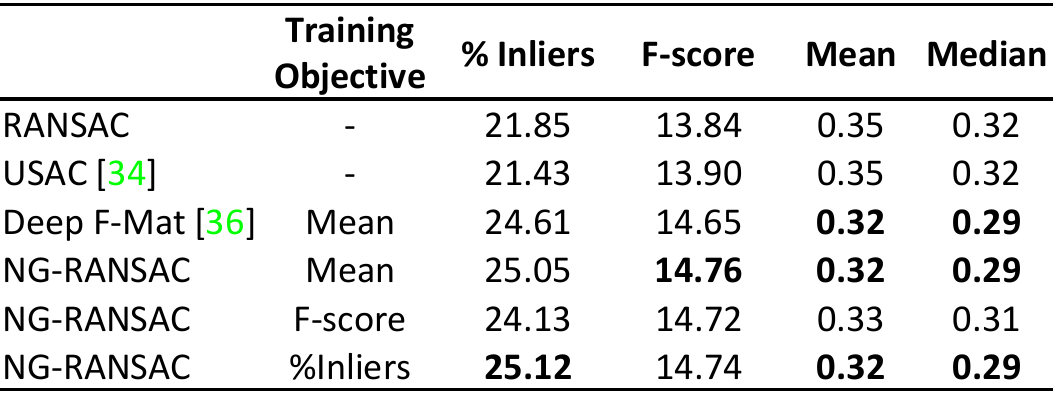}
\end{center}
\vspace{-0.3cm}
   \caption{\textbf{Fundamental Matrix Estimation.} We measure the average percentage of inliers of the estimated model, the alignment of estimated inliers and ground truth inliers (\emph{F-score}), and the mean and median distance of estimated inliers to ground truth epilines. For NG-RANSAC, we compare the performance after training with different objectives. Note that \emph{\%Inliers} is a self-supervised training objective.}
   \vspace{-0.3cm}
\label{fig:table:kitti}
\end{figure}

\noindent \textbf{Datasets.} 
Ranftl and Koltun \cite{deepfund18} evaluate their method on various datasets that involve custom reconstructions not publicly available.
Therefore, we compare to their method on the Kitti dataset \cite{kitti12}, which is online.
Ranftl and Koltun \cite{deepfund18} train their method on sequences 00-05 of the Kitti odometry benchmark, and test on sequences 06-10. 
They form image pairs by taking subsequent images within a sequence.
For each pair, they extract SIFT correspondences and apply Lowe's ratio filter \cite{Lowesift} with a threshold of 0.8.

\noindent \textbf{Evaluation Metric.}
Ranftl and Koltun \cite{deepfund18} evaluate using multiple metrics.
They measure the percentage of inlier correspondences of the final model relative to all correspondences.
They calculate the F-score over correspondences where true positives are inliers of both the ground truth model and the estimated model.
The F-score measures the alignment of estimated and true fundamental matrix in image space.
Both metrics use an inlier threshold of 0.1px.
Finally, they calculate the mean and median epipolar error of inlier correspondences \wrt the ground truth model, using an inlier threshold of 1px.

\noindent \textbf{Implementation}.
We cannot use the architecture of \emph{Deep F-Mat} which is designed for iterative application.
Therefore, we re-use the architecture of Yi \etal \cite{goodcorr18} from the previous section for NG-RANSAC (also see Appendix \ref{app:fmat} for details).
We adhere to the training setup described in Sec.~\ref{sec:exp:emat} with the following changes.
We observed faster training convergence on Kitti, so we omit the initialization stage, and directly optimize the expected task loss (Eq.~\ref{eq:ng-ransac}) for 300k iterations.
Since Ranftl and Koltun \cite{deepfund18} evaluate using multiple metrics, the choice of the task loss function $\loss$ is not clear.
Hence, we train multiple variants with different objectives (\emph{\%Inliers}, \emph{F-score} and \emph{Mean} error) and report the corresponding results.
As minimal solver $f$, we use the 7-point algorithm, a RANSAC threshold of 0.1px, and we draw $K=8$ hypothesis pools per training image with $M=16$ hypotheses each.

\noindent \textbf{Results.}
We report results in Fig.~\ref{fig:table:kitti} where we compare NG-RANSAC with RANSAC, USAC \cite{usac2013} and \emph{Deep F-Mat}.
NG-RANSAC outperforms the classical approaches RANSAC and USAC.
NG-RANSAC also performs slightly superior to \emph{Deep F-Mat}.
We observe that the choice of the training objective has small but significant influence on the evaluation.
All metrics are highly correlated, and optimizing a metric in training generally also achieves good (but not necessarily best) accuracy using this metric at test time.
Interestingly, optimizing the inlier count during training performs competitively, although being a self-supervised objective.
Fig.~\ref{fig:goodcorr2} shows a qualitative result on Kitti.

\begin{figure}[t]
\vspace{-0.3cm}
\begin{center}

   \includegraphics[width=0.9\linewidth]{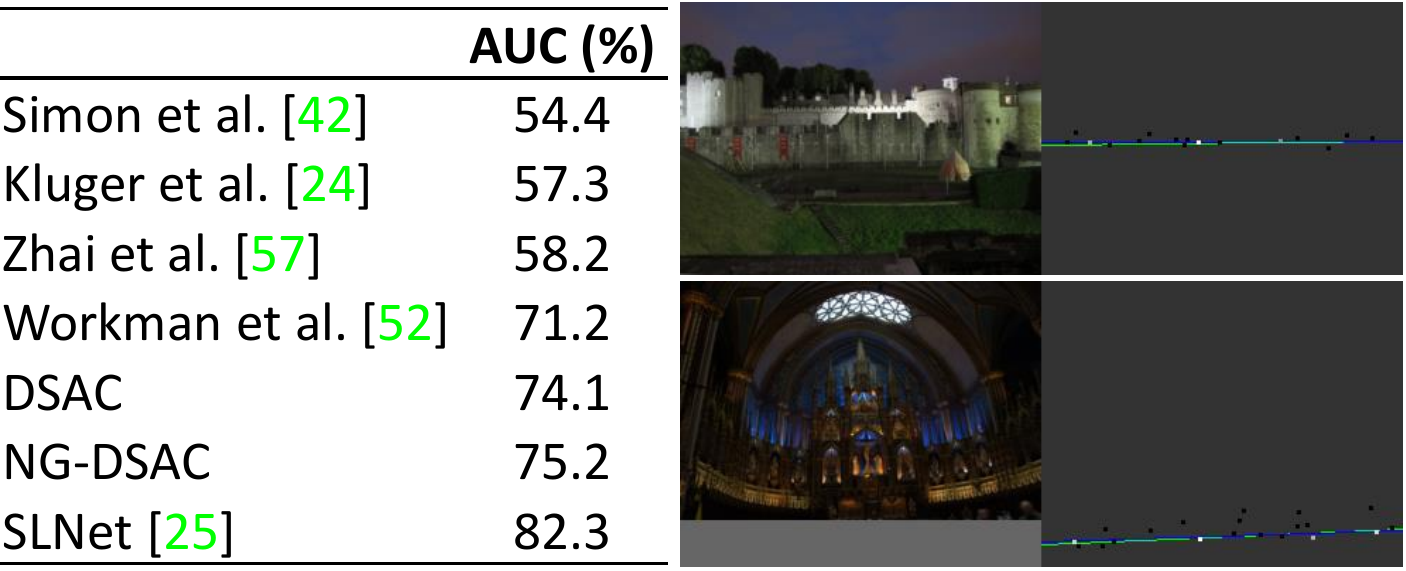}
\end{center}
\vspace{-0.2cm}
   \caption{\textbf{Horizon Line Estimation. Left:} AUC on the HLW dataset. \textbf{Right:} Qualitative results. We draw the ground truth horizon in green and the estimate in blue. Dots mark the observations predicted by NG-DSAC, and the dot colors mark their confidence (dark = low). Note that the horizon can be outside the image.}
\label{fig:horizon}
\vspace{0.14cm}
\end{figure}

\subsection{Horizon Lines}

We fit a parametric model, the horizon line, to a single image.
The horizon can serve as a cue in image understanding \cite{workman2016hlw} or for image editing \cite{SLNet17}.
Traditionally, this task is solved via vanishing point detection and geometric reasoning \cite{Rother2002, KluAck2017, zhai2016horizon, Simon_2018_ECCV}, often assuming a Manhattan or Atlanta world.
We take a simpler approach and use a general purpose CNN that predicts a set of 64 2D points based on the image to which we fit a line with RANSAC, see Fig.~\ref{fig:horizon}.
The network has two output branches predicting (i) the 2D points $\crd(\param) \in \crds(\param)$, and (ii) probabilities $p(\crd;\param)$ for guided sampling (see Appendix \ref{app:hline} for details).

\noindent \textbf{Dataset.} 
We evaluate on the HLW dataset \cite{workman2016hlw} which is a collection of SfM datasets with annotated horizon line.
Test and training images partly show the same scenes, and the horizon line can be outside the image area.

\noindent \textbf{Evaluation Metric.}
As is common practice on HLW, we measure the maximum distance between the estimated horizon and ground truth within the image, normalized by image height.
We calculate the AUC of the cumulative error curve up to a threshold of 0.25.

\begin{figure*}[t!]
\begin{center}
\vspace{-0.3cm}
   \includegraphics[width=1.0\linewidth]{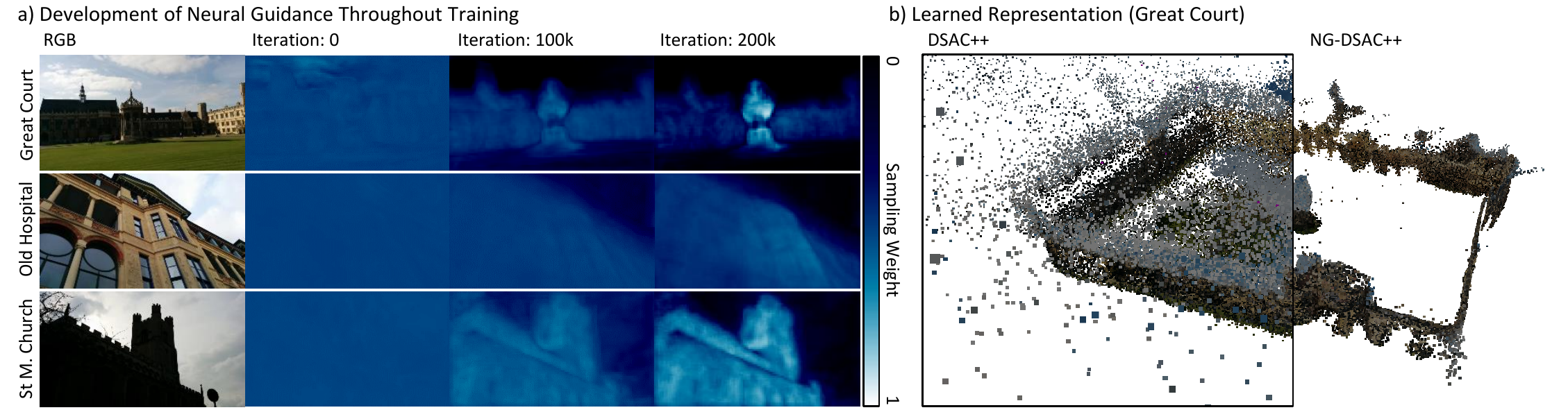}
\end{center}
\vspace{-0.3cm}
   \caption{\textbf{Neural Guidance for Camera Re-localization. a)} Predicted sampling probabilities of NG-DSAC++ throughout training. \textbf{b)} Internal representation of the neural network. We predict scene coordinates for each training image, plotting them with their RGB color. For DSAC++ we choose training pixels randomly, for NG-DSAC++ we choose randomly according to the predicted distribution.}
   \label{fig:camloc}
   \vspace{-0.6cm}
\end{figure*}

\begin{figure}[h!]
\begin{center}
\vspace{0.1cm}
   \includegraphics[width=0.9\linewidth]{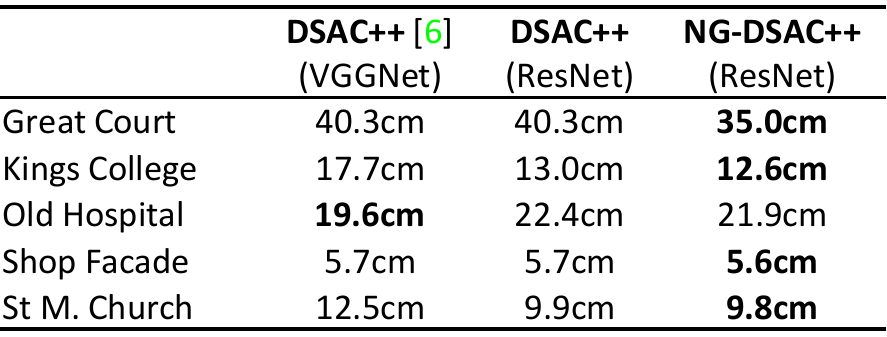}
\end{center}
\vspace{-0.3cm}
   \caption{\textbf{Camera Re-Localization.} We report median position error for Cambridge Landmarks \cite{kendall2015convolutional}. \emph{DSAC++ (ResNet)} is our re-implementation of \cite{brachmann2018lessmore} with an improved network architecture.}
\label{fig:table:camloc} 
 \vspace{-0.5cm}
\end{figure}

\noindent \textbf{Implementation}.
We train using the NG-DSAC objective (Eq.~\ref{eq:ng-dsac}) from scratch for 250k iterations.
As task loss $\loss$, we use the normalized maximum distance between estimated and true horizon.
For hypothesis scoring $s$, we use a soft inlier count \cite{brachmann2018lessmore}.
We train using Adam \cite{adam2014} with a learning rate of 10$^{-4}$. 
For each training image, we draw $K=2$ hypothesis pools with $M=16$ hypotheses. 
We also draw 16 hypotheses at test time.
We compare to DSAC which we train similarly but disable the probability branch.

\noindent \textbf{Results.}
We report results in Fig.~\ref{fig:horizon}.
DSAC and NG-DSAC achieve competitive accuracy on this dataset, ranking among the top methods.
NG-DSAC has a small but significant advantage over DSAC alone.
Our method is only surpassed by SLNet \cite{SLNet17}, an architecture designed to find semantic lines in images.
SLNet generates a large number of random candidate lines, selects a candidate via classification, and refines it with a predicted offset.
We could couple SLNet with neural guidance for informed candidate sampling.
Unfortunately, the code of SLNet is not online and the authors did not respond to inquiries.

\vspace{-0.1cm}
\subsection{Camera Re-Localization}

We estimate the absolute 6D camera pose (position and orientation) \wrt a known scene from a single RGB image. 

\noindent \textbf{Dataset.} 
We evaluate on the Cambridge Landmarks \cite{kendall2015convolutional} dataset.
It is comprised of RGB images depicting five landmark buildings\footnote{We omitted the \emph{Street} scene. Like DSAC++ \cite{brachmann2018lessmore} we failed to achieve sensible results, here. By visual inspection, the corresponding SfM reconstruction seems to be of poor quality, which potentially harms training.} in Cambridge, UK.
Ground truth poses were generated by running a SfM pipeline.

\noindent \textbf{Evaluation Metric.}
We measure the median translational error of estimated poses for each scene\footnote{The median rotational accuracies are between 0.2$^\circ$ to 0.3$^\circ$ for all scenes, and do hardly vary between methods.}.

\noindent \textbf{Implementation}.
We build on the publicly available DSAC++ pipeline \cite{brachmann2018lessmore} which is a scene coordinate regression method \cite{shotton13scorf}.
A neural network predicts for each image pixel a 3D coordinate in scene space.
We recover the pose from the 2D-3D correspondences using a perspective-n-point solver \cite{gao2003complete} within a RANSAC loop.
The DSAC++ pipeline implements geometric pose optimization in a fully differentiable way which facilitates end-to-end training.
We re-implement the neural network integration of DSAC++ with PyTorch (the original uses LUA/Torch).
We also update the network architecture of DSAC++ by using a ResNet \cite{resnet2015} instead of a VGGNet \cite{Simonyan2014}.
As with horizon line estimation, we add a second output branch to the network for estimating a probability distribution over scene coordinate predictions for guided RANSAC sampling.
We denote this extended architecture \emph{NG-DSAC++}.
We adhere to the training procedure and hyperparamters of DSAC++ (see Appendix \ref{app:dsac}) but optimize the NG-DSAC objective (Eq.~\ref{eq:ng-dsac}) during end-to-end training.
As task loss $\loss$, we use the average of the rotational and translational error \wrt the ground truth pose.
We sample $K=2$ hypothesis pools with $M=16$ hypotheses per training image, and increase the number of hypotheses to $M=256$ for testing.

\noindent \textbf{Results}.
We report our quantitative results in Fig.~\ref{fig:table:camloc}.
Firstly, we observe a significant improvement for most scenes when using DSAC++ with a ResNet architecture.
Secondly, comparing DSAC++ with NG-DSAC++, we notice a small to moderate, but consistent, improvement in accuracy.
The advantage of using neural guidance is largest for the \emph{Great Court} scene, which features large ambiguous grass areas, and large areas of sky visible in many images.
NG-DSAC++ learns to ignore such areas, see the visualization in Fig.~\ref{fig:camloc} a).
The network learns to mask these areas solely guided by the task loss during training, as the network fails to predict accurate scene coordinates for them.
In Fig.~\ref{fig:camloc} b), we visualize the internal representation learned by DSAC++ and NG-DSAC++ for one scene.
The representation of DSAC++ is very noisy, as it tries to optimize geometric constrains for sky and grass pixels. 
NG-DSAC++ learns a cleaner representation by focusing entirely on buildings.

\section{Conclusion}

We have presented NG-RANSAC, a robust estimator using guided hypothesis sampling according to learned probabilities.
For training we can incorporate non-differentiable task loss functions and non-differentiable minimal solvers.
Using the inlier count as training objective allows us to also train NG-RANSAC self-supervised.
We applied NG-RANSAC to multiple classic computer vision tasks and observe a consistent improvement \wrt RANSAC alone.

\paragraph*{Acknowledgements:}
This project has received funding from the European Research Council (ERC) under the European Union’s Horizon 2020 research and innovation programme (grant agreement No 647769). The computations were performed on an HPC Cluster at the Center for Information Services and High Performance Computing (ZIH) at TU Dresden.

\appendix

\section{Essential Matrix Estimation}
\label{app:emat}

\noindent \textbf{List of Scenes Used for Training and Testing.}

\vspace{0.2cm}
\noindent Training:
\vspace{0.2cm}
\begin{compactitem}
\item Staint Peter's (Outdoor)
\item brown\_bm\_3 - brown\_bm\_3 (Indoor)
\end{compactitem}

\vspace{0.2cm}
\noindent Testing (Outdoor):
\vspace{0.2cm}
\begin{compactitem}
\item  Buckingham
\item  Notre Dame
\item  Sacre Coeur
\item  Reichstag
\item  Fountain
\item  HerzJesu
\end{compactitem}

\vspace{0.2cm}
\noindent Testing (Indoor):
\vspace{0.2cm}
\begin{compactitem}
\item  brown\_cogsci\_2 - brown\_cogsci\_2
\item  brown\_cogsci\_6 - brown\_cogsci\_6
\item  brown\_cogsci\_8 - brown\_cogsci\_8
\item  brown\_cs\_3 - brown\_cs3
\item  brown\_cs\_7 - brown\_cs7
\item  harvard\_c4 - hv\_c4\_1
\item  harvard\_c10 - hv\_c10\_2
\item  harvard\_corridor\_lounge - hv\_lounge1\_2
\item  harvard\_robotics\_lab - hv\_s1\_2
\item  hotel\_florence\_jx - florence\_hotel\_stair\_room\_all
\item  mit\_32\_g725 - g725\_1
\item  mit\_46\_6conf - bcs\_floor6\_conf\_1
\item  mit\_46\_6lounge - bcs\_floor6\_long
\item  mit\_w85g - g\_0
\item  mit\_w85h - h2\_1
\end{compactitem}

\begin{figure*}
\begin{center}
\includegraphics[width=1.0\linewidth]{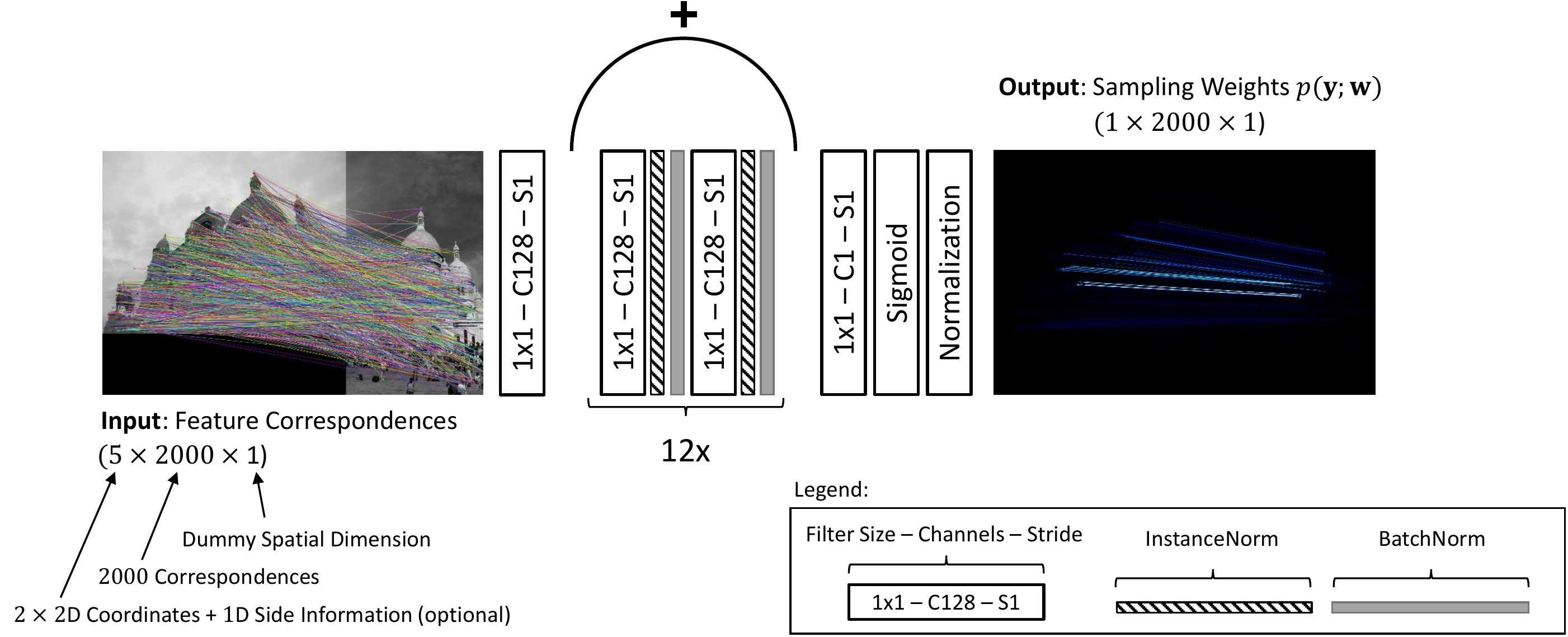}
\end{center}
\vspace{-0.4cm}
   \caption{\textbf{NG-RANSAC Network Architecture for F/E-matrix Estimation.} The network takes a set of feature correspondences as input and predicts as output a weight for each correspondence. The network consists of linear layers interleaved by instance normalization \cite{Ulyanov2016InstanceNT}, batch normalization \cite{batchnorm15} and ReLUs \cite{prelu2015}. The arc with a plus marks a skip connection for each of the twelve blocks \cite{resnet2015}. This architecture was proposed by Yi \etal \cite{goodcorr18}.}
\label{fig:arch:goodcorr}
\vspace{-0.4cm}
\end{figure*}

\noindent \textbf{Network Architecture.}
As mentioned in the main paper, we replicated the architecture of Yi \etal \cite{goodcorr18} for our experiments on epipolar geometry (estimating essential and fundamental matrices).
For a schematic overview see Fig.~\ref{fig:arch:goodcorr}.
The network takes a set of feature correspondences as input, and predicts as output a weight for each correspondence which we use to guide RANSAC hypothesis sampling.
The network consists of a series of multilayer perceptrons (MLPs) that process each correspondence independently.
We implement the MLPs with $1 \times 1$ convolutions.
The network infuses global context via instance normalization layers \cite{Ulyanov2016InstanceNT}, and it accelerate training via batch normalization \cite{batchnorm15}.
The main body of the network is comprised of 12 blocks with skip connections \cite{resnet2015}.
Each block consists of two linear layers followed by instance normalization, batch normalization and a ReLU activation \cite{prelu2015} each.
We apply a Sigmoid activation to the last layer, and normalize by dividing by the sum of outputs.\footnote{The original architecture of  Yi \etal \cite{goodcorr18} uses a slightly different output processing due to using the output as weights for a robust model fit. They use a ReLU activation followed by a tanh activation.}

\noindent \textbf{Initialization Procedure}.
We initialize our network in the following way.
We define a target sampling distribution $g(\crd; \gt{E})$ using the ground truth essential matrix $\gt{E}$ given for each training pair.
Intuitively, the target distribution should return a high probability when a correspondence $\crd$ is aligned with the ground truth essential matrix $\gt{E}$, and a low probability otherwise.
We assume that correspondence $\crd$ is a 4D vector containing two 2D image coordinates $\mathbf{x}$ and $\mathbf{x'}$ (3D in homogeneous coordinates).
We define the epipolar error of a correspondence \wrt essential matrix $E$:
\begin{equation}
d(\crd, E) = \frac{(\mathbf{x'}^\top E \mathbf{x})^2}{[E \mathbf{x}]_0^2+[E \mathbf{x}]_1^2+[E^\top \mathbf{x}']_0^2+[E^\top \mathbf{x}']_1^2},
\end{equation}
where $[\cdot]_i$ returns the $i$th entry of a vector.
Using the epipolar error, we define the target sampling distribution:
\begin{equation}
g(\crd;\gt{E}) = \frac{1}{2 \pi \sigma^2} \exp \left(- \frac{d(\crd, \gt{E})}{2 \sigma^2} \right).
\end{equation}
Parameter $\sigma$ controls the softness of the target distribution, and we use $\sigma=10^{-3}$ which corresponds to the inlier threshold we use for RANSAC.
To initialize our network, we minimize the KL divergence between the network prediction $p(\crd; w)$ and the target distribution $g(\crd;\gt{E})$.
We initialize for 75k iterations using Adam \cite{adam2014} with a learning rate of $10^{-3}$ and a batch size of 32.

\noindent \textbf{Implementation Details}.
For the following components we rely on the implementations provided by OpenCV \cite{opencv_library}: the 5-point algorithm \cite{fivepoint04}, epipolar error, SIFT features \cite{Lowesift}, feature matching, and essential matrix decomposition.
We extract 2000 features per input image which yields 2000 correspondences for image pairs after matching.
When applying Lowe's ratio criterion \cite{Lowesift} for filtering and hence reducing the number of correspondences, we randomly duplicate correspondences to restore the number of 2000.
We minimize the expected task loss using Adam \cite{adam2014} with a learning rate of $10^{-5}$ and a batch size of 32. 
We choose hyperparameters based on validation error of the \emph{Reichstag} scene.
We observe that the magnitude of the validation error corresponds well to the magnitude of the training error, \ie a validation set would not be strictly required.

When calculating the AUC for evaluation, we adhere to the protocol of Yi et al.~\cite{goodcorr18} to ensure comparability.
Yi et al.~approximate the AUC via the area under the cumulative histogram with a bin width of 5$^\circ$.

\begin{figure*}
\begin{center}
\includegraphics[width=0.8\linewidth]{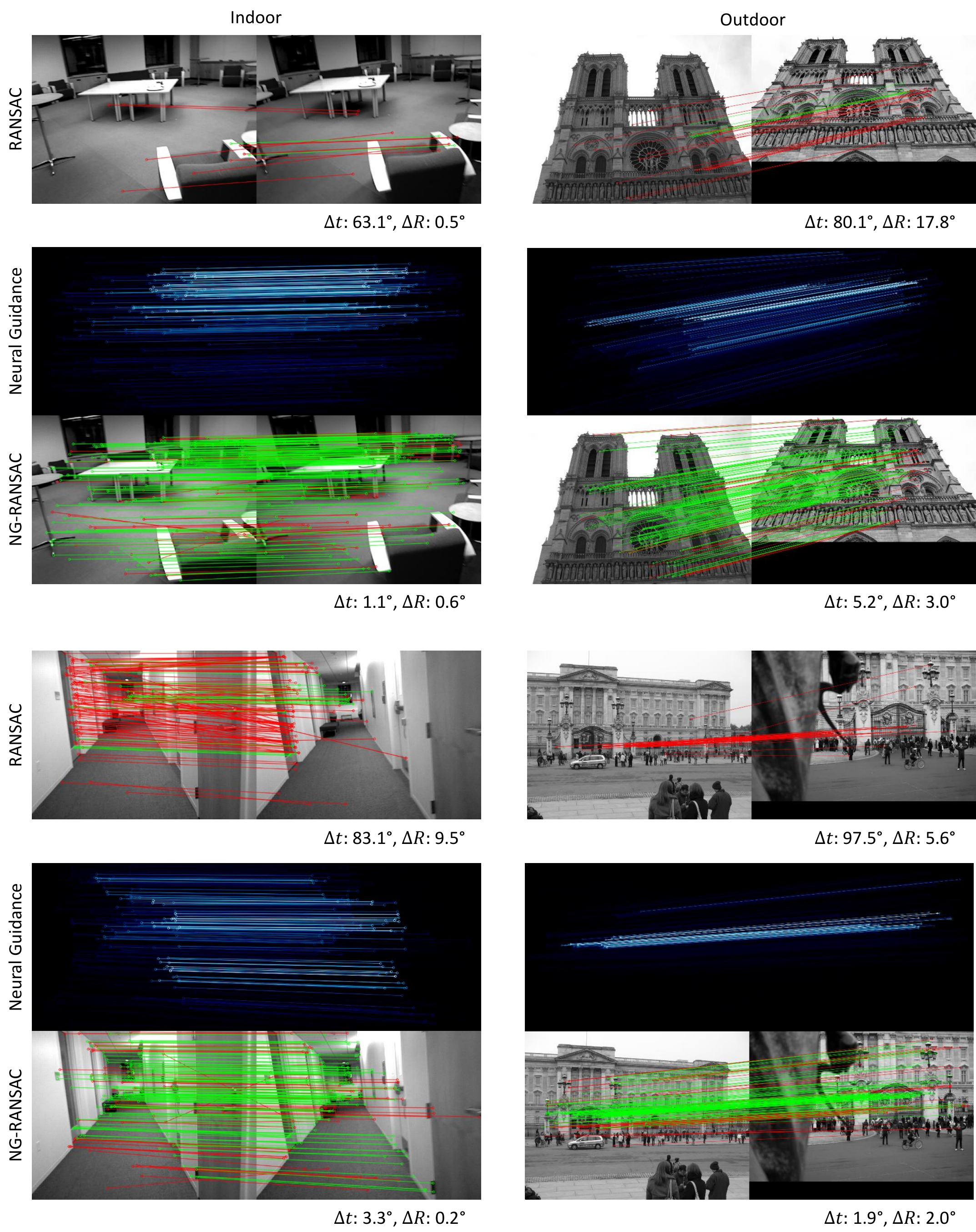}
\end{center}
   \vspace{-0.3cm}
   \caption{\textbf{Qualitative Results for Essential Matrix Estimation.} We compare results of RANSAC and NG-RANSAC. Below each result, we give the angular error between estimated and true translation vectors, and estimated and true rotation matrices. We draw correspondences in green if they adhere to the ground truth essential matrix with an inlier threshold of $10^{-3}$, and red otherwise.}
   \vspace{-0.4cm}
\label{fig:results:goodcorr}

\end{figure*}

\noindent \textbf{Qualitative Results.}
We present additional qualitative results for indoor and outdoor scenarios in Fig.~\ref{fig:results:goodcorr}.
We compare results of RANSAC and NG-RANSAC, also visualizing neural guidance as predicted by our network.
We obtain these results in the high-outlier setup, \ie without using Lowe's ratio criterion and without using side information as additional network input.

\vspace{-0.1cm}
\section{Fundamental Matrix Estimation}
\label{app:fmat}

\noindent \textbf{Implementation Details.}
We reuse the architecture of Fig.~\ref{fig:arch:goodcorr}.
To normalize image coordinates of feature matches, we subtract the mean coordinate and divide by the coordinate standard deviation, where we calculate mean and standard deviation over the training set.
Ranftl and Koltun \cite{deepfund18} fit the final fundamental matrix to the top 20 weighted correspondences as predicted by their network.
Similarly, we re-fit the final fundamental matrix to the largest inlier set found by NG-RANSAC.
This refinement step results in a small but noticeable increase in accuracy.
For the following components we rely on the implementations provided by OpenCV \cite{opencv_library}: the 7-point algorithm, epipolar error, SIFT features \cite{Lowesift} and feature matching.

\begin{figure*}
\begin{center}
\includegraphics[width=0.74\linewidth]{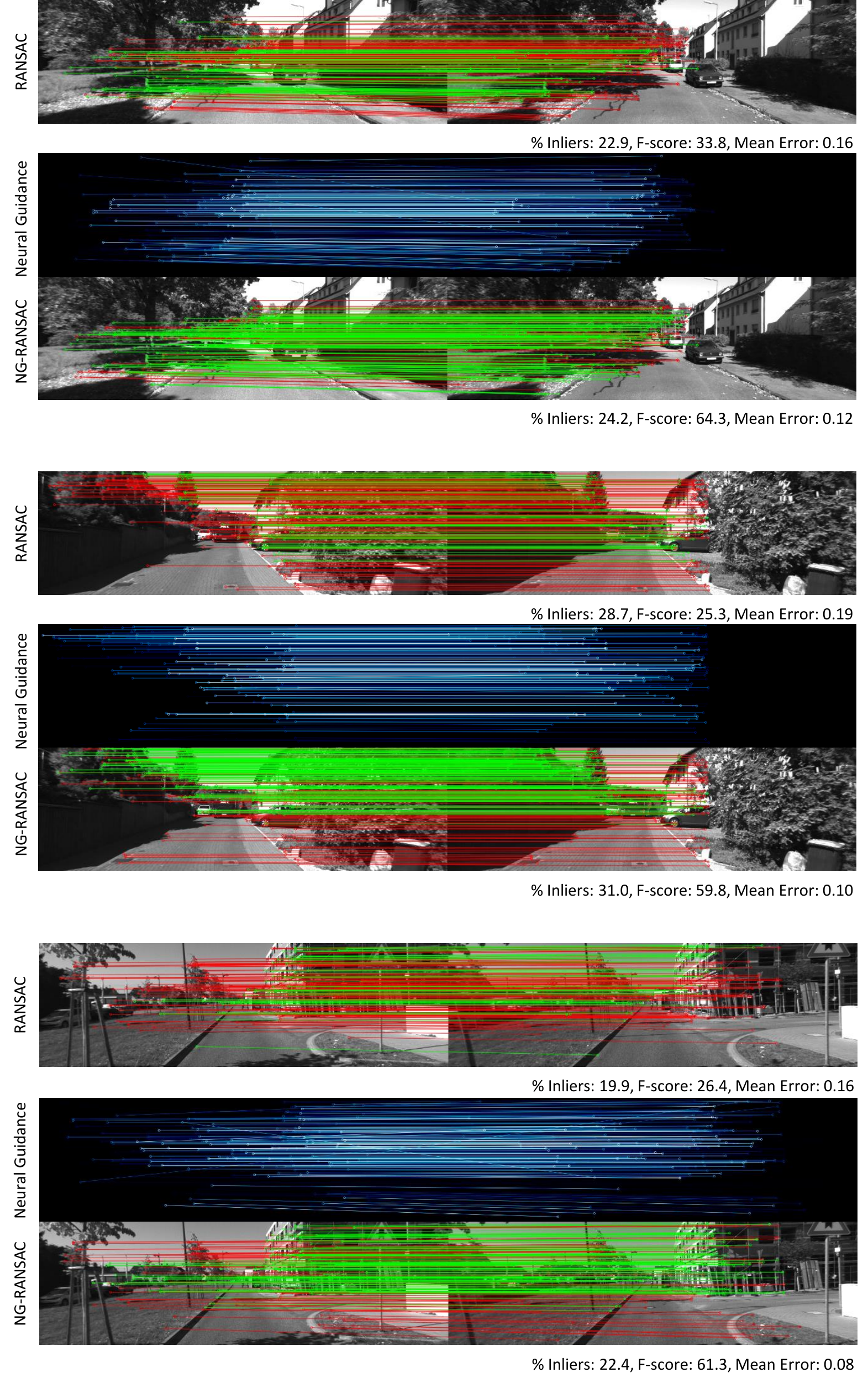}
\end{center}
   \caption{\textbf{Qualitative Results for Fundamental Matrix Estimation.} We compare results of RANSAC and NG-RANSAC. Below each result, we give the percentage of inliers of the final model, the F-score which measures the alignment of estimated and true fundamental matrix, and the mean epipolar error of estimated inlier correspondences \wrt the ground truth fundamental matrix. We draw correspondences in green if they adhere to the ground truth fundamental matrix with an inlier threshold of $0.1$px, and red otherwise.}
   \vspace{-0.5cm}
\label{fig:results:kitti}
\end{figure*}

\noindent \textbf{Qualitative Results.}
We present additional qualitative results for the Kitti dataset \cite{kitti12} in Fig.~\ref{fig:results:kitti}.
We compare results of RANSAC and NG-RANSAC, also visualizing neural guidance as predicted by our network.

\vspace{-0.2cm}
\section{Horizon Lines}
\vspace{-0.1cm}
\label{app:hline}

\begin{figure*}
\begin{center}
\includegraphics[width=1.0\linewidth]{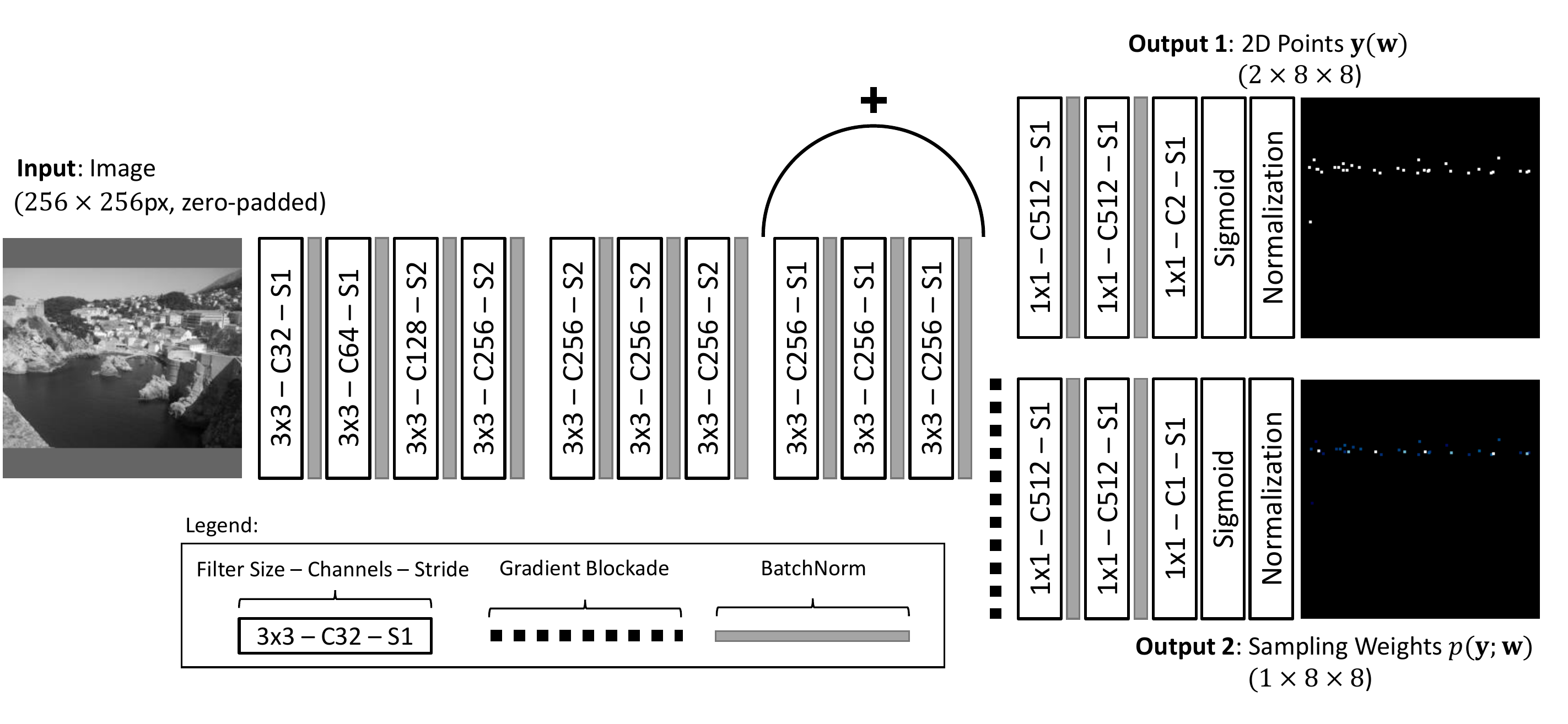}
\end{center}
\vspace{-0.5cm}
   \caption{\textbf{NG-DSAC Network Architecture for Horizon Line Estimation.} The network takes a grayscale image as input and predicts as output a set of 2D points and corresponding sampling weights. The network consists of convolution layers interleaved by batch normalization \cite{batchnorm15} and ReLUs \cite{prelu2015}. The arc with a plus marks a skip connection \cite{resnet2015}. We use the gradient blockage during training to prevent direct influence of the sampling prediction (second branch) to learning the observations (first branch).}
\label{fig:arch:horizon}
   \vspace{-0.3cm}
\end{figure*}

\noindent \textbf{Network Architecture.}
We provide a schematic of our network architecture for horizon line estimation in Fig.~\ref{fig:arch:horizon}.
The network takes a $256 \times 256$px image as input.
We re-scale images of arbitrary aspect ratio such that the long side is $256$px.
We symmetrically zero-pad the short side to $256$px.
The network has two output branches. 
The first branch predicts a set of $8 \times 8 = 64$ 2D points, our observations $\crd(\param)$, to which we fit the horizon line.
We apply a Sigmoid and re-scale output points to [-1.5,1.5] in relative image coordinates to support horizon lines outside the image area.
We implement the network in a fully convolutional way \cite{fcn2015}, \ie each output point is predicted for a patch, or restricted receptive field, of the input image.
Therefore, we shift the coordinate of each output point to the center of its associated patch.

The second branch predicts sampling probabilities $p(\crd; \param)$ for each output point.
We apply a Sigmoid to the output of the second branch, and normalize by dividing by the sum of outputs.
During training, we block the gradients of the second output branch when back propagating to the base network.
The sampling gradients have larger variance and magnitude than the observation gradients of the first branch, especially in the beginning of training with a negative effect on convergence of the network as a whole.
Intuitively, we want to give priority to the observation prediction because they determine the accuracy of the final model parameters.
The sampling prediction should address deficiencies in the observation predictions without influencing them too much.
The gradient blockade ensures these properties.

\noindent \textbf{Implementation Details.}
We use a differentiable soft inlier count \cite{brachmann2018lessmore} as scoring function, \ie:
\begin{equation}
s(\mdl, \crds) = \alpha \sum_{\crd \in \crds} 1 - \text{sig}[\beta d(\crd, \mdl) - \beta \tau],
\end{equation}
where $d(\crd, \mdl)$ denotes the point-line distance between observation $\crd$ and line hypothesis $\mdl$.
Hyperparameter $\alpha$ determines the softness of the scoring distribution in DSAC, $\beta$ determines the softness of the Sigmoid, and $\tau$ is the inlier threshold.
We use $\alpha = 0.1$, $\beta = 100$ and $\tau = 0.05$.

We convert input images to grayscale, and apply the following data augmentation strategy during training:
We randomly adjust brightness and contrast in the range of $\pm 10\%$.
We randomly rotate/scale/shift images (and ground truth horizon lines) in the range of $\pm 5^\circ$/$20\%$/$8$px.

As discussed in the main paper, we use the normalized maximum distance between a line hypothesis and the ground truth horizon in the image as task loss $\loss$.
This can lead to stability issues when we sample line hypotheses with very steep slope.
Therefore, we clamp the task loss to a maximum of 1, \ie the normalized image height.

As mentioned before, some images in the HLW dataset \cite{workman2016hlw} have their horizon outside the image. 
Some of these images contain virtually no visual cue where the horizon exactly lies.
Therefore, we find it beneficial to use a robust variant of the task loss $\loss'$ that limits the influence of such outliers.
We use:
\begin{equation}
\loss' = \begin{cases}
\loss & \loss < 0.25 \\
0.25 \sqrt{\loss} & \text{otherwise} \\
\end{cases},
\end{equation}
\ie we use the square root of the task loss after a magnitude of $0.25$, which is the magnitude up to which the AUC is calculated when evaluating on HLW \cite{workman2016hlw}.

\begin{figure*}
\begin{center}
\includegraphics[width=0.9\linewidth]{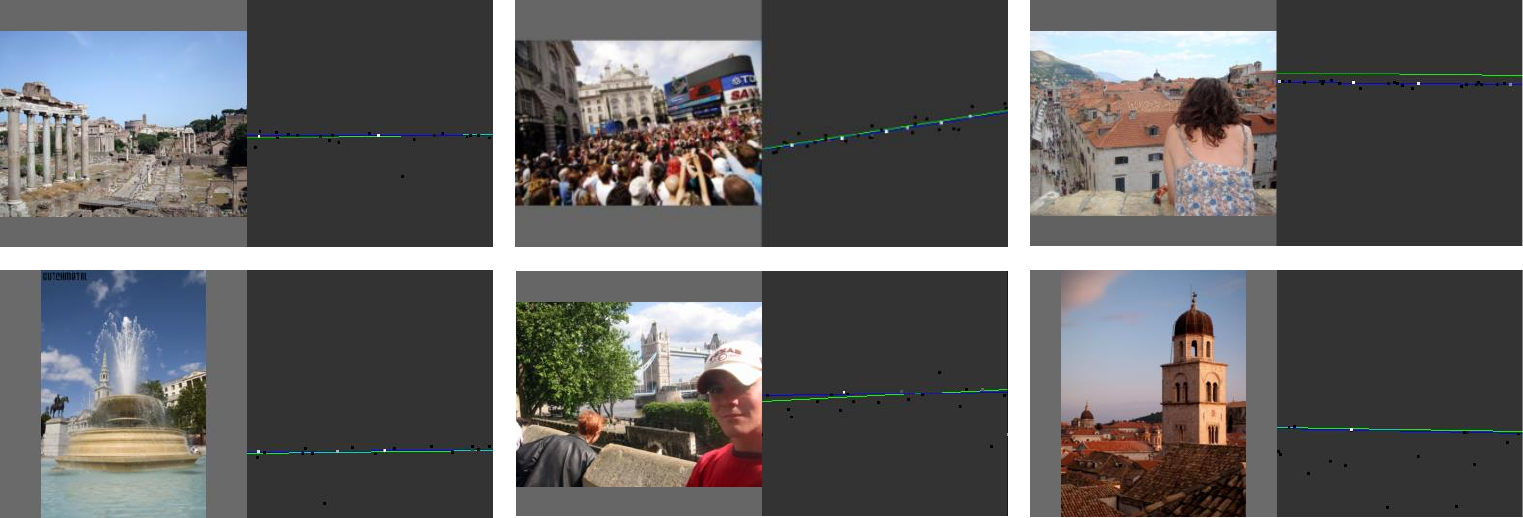}
\end{center}
\vspace{-0.3cm}
   \caption{\textbf{Qualitative Results for Horizon Line Estimation.} Next to each input image, we show the estimated horizon line in blue and the true horizon line in green. We also show the observation points predicted by our network, colored by their sampling weight (dark = low).}
   
\label{fig:results:horizon}
\end{figure*}

\noindent \textbf{Qualitative Results.}
We present additional qualitative results for the HLW dataset \cite{workman2016hlw} in Fig.~\ref{fig:results:horizon}.

\section{Camera Re-Localization}
\label{app:dsac}

\begin{figure*}
\begin{center}
\includegraphics[width=0.9\linewidth]{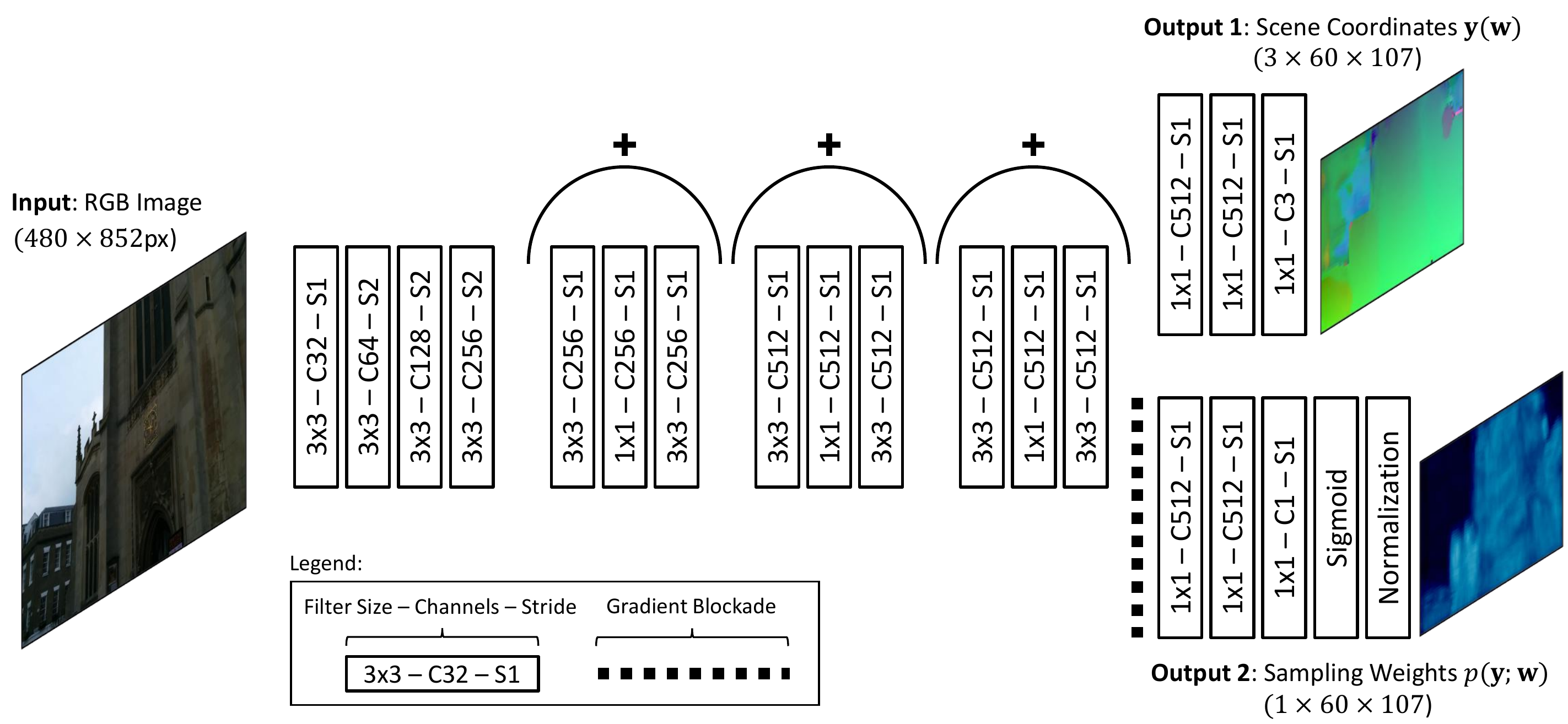}
\end{center}
   \caption{\textbf{NG-DSAC++ Network Architecture for Camera Re-Localization.} The network takes an RGB image as input and predicts as output dense scene coordinates and corresponding sampling weights. The network consists of convolution layers followed by ReLUs \cite{prelu2015}. Am arc with a plus marks a skip connection \cite{resnet2015}. We use the gradient blockage during training to prevent direct influence of the sampling prediction (second branch) to learning the scene coordinates (first branch).}
\label{fig:arch:camloc}
\end{figure*}

\begin{figure*}[t!]
\begin{center}
\includegraphics[width=0.9\linewidth]{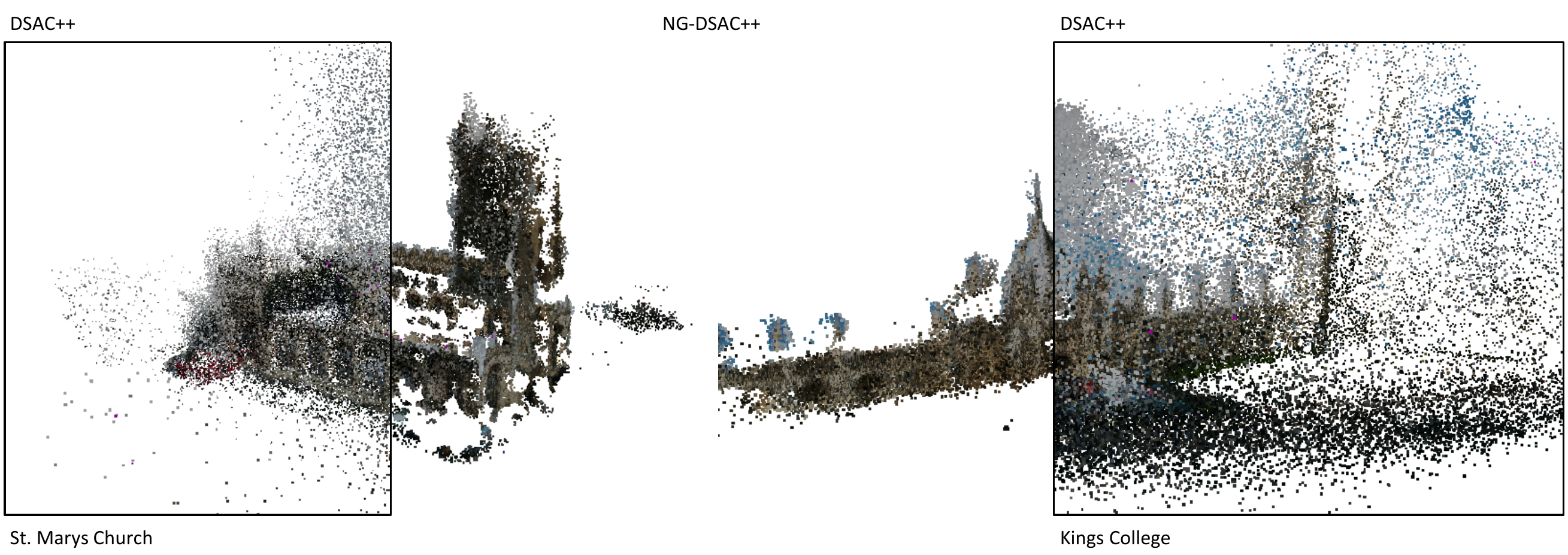}
\end{center}
\vspace{-0.3cm}
   \caption{\textbf{Learned 3D Representations.} We visualize the internal representation of the neural network. We predict scene coordinates for each training image, plotting them with their RGB color. For DSAC++ we choose training pixels randomly, for NG-DSAC++ we choose randomly among the top 1000 pixels per training image according to the predicted distribution.}
   \vspace{-0.3cm}
\label{fig:results:camloc}
\end{figure*}

\noindent \textbf{Network Architecture.}
We provide a schematic of our network architecture for camera re-localization in Fig.~\ref{fig:arch:camloc}.
The network is a FCN \cite{fcn2015} that takes an RGB image as input, and predicts dense outputs, sub-sampled by a factor of 8.
The network has two output branches. 
The first branch predicts 3D scene coordinates \cite{shotton13scorf}, our observations $\crd(\param)$, to which we fit the 6D camera pose.
The second output branch predicts sampling probabilities $p(\crd; \param)$ for the scene coordinates.
We apply a Sigmoid to the output of the second branch, and normalize by dividing by the sum of outputs.
During training, we block the gradients of the second output branch when back propagating to the base network.
The sampling gradients have larger variance and magnitude than the observation gradients of the first branch, especially in the beginning of training.
This has a negative effect on convergence of the network as a whole.
Intuitively, we want to give priority to the scene coordinate prediction because they determine the accuracy of the pose estimate.
The sampling prediction should address deficiencies in the scene coordinate predictions without influencing them too much.
The gradient blockade ensures these properties.

\noindent \textbf{Implementation details.}
We follow the three-stage training procedure proposed by Brachmann and Rother for DSAC++ \cite{brachmann2018lessmore}.

Firstly, we optimize the distance between predicted and ground truth scene coordinates.
We obtain ground truth scene coordinates by rendering the sparse reconstructions given in the Cambridge Landmarks dataset \cite{kendall2015convolutional}.
We ignore pixels with no corresponding 3D point in the reconstruction.
Since the reconstructions contain outlier 3D points, we use the following robust distance:
\begin{equation}
\d(\crd, \gt{\crd}) = \begin{cases}
||\crd - \gt{\crd}||_2 & ||\crd - \gt{\crd}||_2 < 10 \\
10 \sqrt{||\crd - \gt{\crd}||_2} & \text{otherwise}
\end{cases},
\end{equation}
\ie we use the Euclidean distance up to a threshold of 10m after which we use the square root of the Euclidean distance.
We train the first stage for 500k iterations using Adam \cite{adam2014} with a learning rate of $10^{-4}$ and a batch size of 1 image.

Secondly, we optimize the reprojection error of the scene coordinate predictions \wrt to the ground truth camera pose.
Similar to the first stage, we use a robust distance function with a threshold of 10px after which we use the square root of the reprojection error.
We train the second stage for 300k iterations using Adam \cite{adam2014} with a learning rate of $10^{-4}$ and a batch size of 1 image.

Thirdly, we optimize the expected task loss according to the NG-DSAC objective as explained in the main paper.
As task loss we use $\loss = \angle(\boldsymbol{\theta}, \gt{\boldsymbol{\theta}}) + ||\mathbf{t} - \gt{\mathbf{t}}||_2$.
We measure the angle between estimated camera rotation $\boldsymbol{\theta}$ and ground truth rotation $\gt{\boldsymbol{\theta}}$ in degree.
We measure the distance between the estimated camera position $\mathbf{t}$ and ground truth position $\gt{\mathbf{t}}$ in meters.
As with horizon line estimation (see previous section), we use a soft inlier count as hypothesis scoring function with hyperparameters $\alpha = 10$, $\beta = 0.5$ and $\tau = 10$.
We train the third stage for 200k iterations using Adam \cite{adam2014} with a learning rate of $10^{-6}$ and a batch size of 1 image.

\noindent \textbf{Learned 3D Representations.}
We visualize the internal 3D scene representations learned by DSAC++ and NG-DSAC++ in Fig.~\ref{fig:results:camloc} for two more scenes.

{\small
\bibliographystyle{ieee}
\bibliography{ngransac}
}
\end{document}